\documentclass[preprint,12pt,authoryear]{elsarticle}

\usepackage{amssymb}
\usepackage[authoryear]{natbib}
\usepackage{subcaption}

\usepackage{tikz}
\usepackage{float}
\usepackage[edges]{forest}
\usepackage[T1]{fontenc}
\usepackage{lmodern}   
\usepackage{multirow}
\usepackage{booktabs}
\usepackage{amsmath}
\usepackage{url}
\usepackage{gensymb}
\usepackage{graphicx}
\usepackage{verbatim}
\usepackage{tabularx} 
\usepackage{algorithm}
\usepackage{algorithmicx}
\usepackage{algpseudocode}
\usepackage{minted}
\usepackage{listings}
\usepackage[utf8]{inputenc}
\RequirePackage{stfloats}
\usepackage{xurl}
\usepackage[colorlinks=true, linkcolor=blue, citecolor=blue, urlcolor=blue, hidelinks]{hyperref} 
\usepackage{adjustbox}
\usepackage{xcolor}

\def\tsc#1{\csdef{#1}{\textsc{\lowercase{#1}}\xspace}}
\tsc{WGM}
\tsc{QE}

\begin{document}

\begin{frontmatter}


\title{A ROS2 Benchmarking Framework for Hierarchical Control Strategies in Mobile Robots for Mediterranean Greenhouses}


\author[UAL-Inf]{Fernando Cañadas-Aránega}\corref{corr}\ead{fernando.ca@ual.es}
\cortext[corr]{Corresponding author}

\author[UNED]{Francisco José Mañas-Álvarez}\ead{fjmanas@dia.uned.es}

\author[UAL-Inf]{José Luis Guzmán}\ead{joguzman@ual.es}

\author[UAL-Inf]{José C. Moreno}\ead{jcmoreno@ual.es}

\author[UAL-Eng]{José L. Blanco-Claraco}\ead{jlblanco@ual.es}


\affiliation[UAL-Inf]{
    organization={Department of Informatics, CIESOL, ceiA3, Universidad de Almería},
    addressline={Ctra. Sacramento s/n},
    city={Almería},
    postcode={04120},
    country={Spain}
}

\affiliation[UNED]{
    organization={Department of Computer Science and Automation, National University of Distance Education (UNED)},
    addressline={C/ Juan del Rosal 16},
    city={Madrid},
    postcode={28040},
    country={Spain}
}

\affiliation[UAL-Eng]{
    organization={Department of Engineering, CIESOL, ceiA3, Universidad de Almería},
    addressline={Ctra. Sacramento s/n},
    city={Almería},
    postcode={04120},
    country={Spain}
}


\begin{abstract}
Mobile robots operating in agro-industrial environments, such as Mediterranean greenhouses, are subject to challenging conditions including uneven terrain, variable friction, payload changes, and terrain slopes, all of which significantly affect control performance and stability. Despite the increasing adoption of robotic platforms in agriculture, the lack of standardized and reproducible benchmarks hinders fair comparison and systematic evaluation of control strategies under realistic operating conditions. This paper presents a comprehensive benchmarking framework for the evaluation of mobile robot controllers in greenhouse environments. The proposed framework integrates an accurate three-dimensional model of the environment, a physics-based simulator, and a hierarchical control architecture comprising low-, mid-, and high-level control layers. Three benchmark categories are defined to enable modular assessment ranging from actuator-level control to full autonomous navigation. Additionally, three disturbance scenarios—payload variation, terrain type, and slope—are explicitly modeled to replicate real-world agricultural conditions. To ensure objective and reproducible evaluation, standardized performance metrics are introduced, including the Squared Absolute Error (SAE), the Squared Control Input (SCI), and composite performance indices. Statistical analysis based on repeated trials is employed to mitigate the influence of sensor noise and environmental variability. The framework is further enhanced by a plugin-based architecture that facilitates seamless integration of user-defined controllers and planners. The proposed benchmark provides a robust and extensible tool for the quantitative comparison of classical, predictive, and planning-based control strategies in realistic conditions, bridging the gap between simulation-based analysis and real-world agro-industrial applications.
\end{abstract}



\end{frontmatter}

\section{Introduction} \label{1. Intrducciotion}
The greenhouse agriculture sector has become consolidated over recent decades as a fundamental strategy for improving food productivity and optimizing the use of water and energy resources. Part of these advances is due to the progressive integration of automation of systems and advanced control mechanisms in key processes, such as temperature, humidity, and ventilation management \citep{van2010optimal}. In the field of automation, mobile robots designed for transportation, inspection, and operator-assistance tasks have emerged as essential actors in the transition toward digitalized and connected agriculture \citep{hernandez2025reconfigurable}. Although greenhouses constitute structured environments, they present relevant challenges for autonomous mobility: narrow corridors that optimize cultivation but limit maneuverability; terrains with nonlinear behaviors—including slippage, heterogeneous compaction, or the presence of plant residues—dynamic obstacles, and changing environmental conditions \citep{ryu2024evaluation}. These factors hinder navigation and require robust control strategies capable of ensuring precise and safe motions even under uncertainty.

From a control perspective, strategies applied to mobile robots range from classical approaches such as Proportional-Integral-Derivative (PID) controller \citep{rosero2023smart}, to more sophisticated approaches based on Predictive Control \citep{kraus2013moving}, adaptive control \citep{liu2023novel}, or data-driven and machine-learning-based methods \citep{yang2022intelligent}. However, many works address only one of the levels of the robotic chain — perception/ localization, planning, trajectory tracking, or actuator control—or rely on overly idealized scenarios, with minimal disturbances or without reproducibility, which complicates objective comparison and the identification of strengths and limitations between classical and emerging methods. To overcome these limitations, testbeds (\textit{benchmarks} platforms) are essential tools: they allow the replication of real conditions, as disturbances and other uncertainties, and guarantee the repeatability of experiments.

In this regard, robotic simulators have become indispensable tools for the development and validation of mobile robots. Widely adopted platforms such as Gazebo \citep{koenig2004design}, Webots \citep{michel2004cyberbotics}, or CoppeliaSim \citep{rohmer2013v}, have proved highly valuable for navigation, control, and perception tasks, particularly in service robotics and urban environments. Nevertheless, their ability to reproduce phenomena characteristic of agricultural environments, such as nonlinear soil behavior, geometries that change due to plant growth, residue accumulation, or motion constraints imposed by narrow corridors, is still limited. As a result, the rigorous evaluation of control algorithms becomes challenging when the goal is to capture the real complexity of greenhouse environments \citep{shamshiri2018simulation}.

In recent years, the MVSim has emerged as a lightweight and computationally efficient alternative for multivehicle ground simulation, providing physically consistent wheel--soil interaction models and advanced support for multi-agent real-time testing \citep{blanco2023multivehicle}. Its modular architecture makes it particularly suitable for benchmarking advanced control strategies, and its applicability to agricultural scenarios—such as Ackermann-steered robots operating in Mediterranean greenhouses—has already been demonstrated \citep{aranega2024navegacion}. Furthermore, its integration with ROS/ROS2 and compatibility with IoT technologies through FIWARE \citep{munoz2020new} facilitate end-to-end experimentation, enabling the evaluation of complete control pipelines rather than isolated algorithmic components. These features position MVSim not merely as a simulation tool, but as a solid backbone for systematic and reproducible benchmarking in agricultural robotics.

From the viewpoint of control methodologies, the literature reports a broad spectrum of approaches designed to cope with nonlinearities, uncertainties, and environmental variability. PID controllers remain widely used due to their simplicity and robustness, and several extensions—including feedforward compensation—have demonstrated improved performance on terrains with variable friction and deformable soils \citep{canadas2024pid}. Their versatility enables deployment across differential, omnidirectional, and Ackermann-based mobile platforms. Adaptive control strategies have also become increasingly relevant in scenarios involving rapidly changing dynamics, such as soils with fluctuating moisture or compaction, where online parameter estimation has shown consistent trajectory-tracking performance under unstructured conditions \citep{fasiolo2023towards}. In parallel, optimal control approaches, particularly Model Predictive Control (MPC), have attracted increasing attention due to their ability to explicitly handle nonlinear dynamics, constraints, and multivariable interactions \citep{bhat2025revisiting}. Hybrid and intelligent control techniques, including neural networks and fuzzy control, are likewise emerging as promising alternatives capable of adapting to partially observable and time-varying agricultural conditions \citep{alrowaily2024application}. Nevertheless, the performance of such methods is strongly influenced by the fidelity of the simulation environments used for design and validation, which is critical for reliable sim-to-real transfer.

Benchmarking efforts have proven instrumental in enabling systematic comparison of control and navigation strategies across multiple domains. Several initiatives have introduced standardized scenarios, models, and performance metrics to facilitate reproducible evaluation \citep{yan2025benchmarking}. For instance, \citep{song2022towards} proposes a benchmark focused on model identification for UAVs with variable dynamics, while \citep{menghini2024modelling} provides a reference model for the design and tuning of controllers in underwater vehicles (UUVs). In the aerospace domain, the \textit{European Satellite Benchmark} \citep{sanfedino2024european} illustrates how open benchmarks support the comparative assessment of advanced control strategies for both research and education.

In mobile robotics, benchmarks focused on navigation, SLAM, and trajectory planning---usable both in simulators and via high-quality datasets---are particularly widespread. In structured ground-navigation contexts, \citep{perille2020benchmarking} presents a benchmark for metric navigation designed to evaluate controller performance in static and moderately structured environments. For more dynamic scenarios, Arena-Bench \citep{kastner2022arena} offers a testbed for comparing obstacle-avoidance approaches in highly dynamic environments. Likewise, \citep{heiden2021bench} introduces \textit{Bench-MR}, a motion-planning benchmark for wheeled robots subject to kinematic constraints. More recently, Verti-Arena \citep{chen2025verti} provides a controllable testbed for off-road autonomy across mixed terrains, illustrating the importance of standardized physical scenarios with controlled variability. In the agricultural domain, \citep{kulathunga2025navigating} proposes a comprehensive framework specifically aimed at navigation in narrow agricultural corridors, highlighting the growing interest in benchmarks tailored to representative agricultural conditions.

However, none of the existing benchmarks explicitly addresses the combination of (i) soil dynamics specific to agricultural environments, (ii) the geometry of Mediterranean greenhouses, (iii) robot--crop interaction, and (iv) the full hierarchy of control levels---from actuator-level control to autonomous navigation. The absence of tools enabling the systematic evaluation of diverse controllers under reproducible conditions remains a well-recognized gap in the community. Consequently, there is a growing need for a modular, open, and physically representative benchmark that supports systematic comparison of different strategies of control. 

To address this need, this work presents an integrated \textit{benchmark} designed to evaluate control strategies for mobile robots operating in an agrotechnological greenhouse environment. The platform is based on MVSim and integrates an accurate 3D model of the real Agroconnect facilities (Almería, Spain), generated from designs and data acquired in real operations \citep{canadas2024multimodal}. The proposal offers four main contributions:
(i) a detailed 3D environment, together with a physical model of soil–wheel interaction calibrated with parameters of the real greenhouse;
(ii) a modular control architecture—\textit{low-} / \textit{mid-} / \textit{high-} level—that allows comparison between classical and advanced methodologies (PID, gain scheduling, Model Predictive Control (MPC), Model Predictive Path Integral (MPPI) control, adaptive control, learning-based strategies, etc.);
(iii) a set of standardized scenarios and metrics—trajectory error, mission time, collisions, simulated energy consumption, resilience to faults—that facilitate objective and reproducible evaluation of each controller’s performance;
and (iv) open-source publication, in order to encourage reproducibility, methodological comparability, and extension by the community.
This \textit{benchmark} aims to become a reference for researchers and developers, enabling the replication of complex scenarios, the analysis of robot interaction with real greenhouse elements, and the rigorous comparison of different control strategies. Thus, the platform contributes to a transparent and reproducible evaluation of mobile robot behavior in protected agriculture.

In Section \ref{sec: 3. Enviroment} describes the experimental environment, including the simulator employed, as well as the characteristics of the map and the robot used. Section \ref{sec: 4. Low level} introduces the proposed control schemes together with their mathematical analysis, structured into low-, mid-, and high-level layers that enable different degrees of interaction for the user. In Section \ref{sec: 5. Competicion}, all aspects of the benchmark challenge are described, including a reference case that users may adopt as a baseline for improvement. Finally, Section \ref{sec: 8. Conclusion} presents the main conclusions derived from this work.
\section{Environment setup} \label{sec: 3. Enviroment}

This section describes the elements that make up the simulator, performing a mathematical analysis of the proposed systems and diagrams.

\subsection{Experimental configuration} \label{3.1. Models}

This subsection summarizes the 3D models used in the benchmark as well as the simulation environment.

\subsubsection{Greenhouse model} \label{3.1.1. Invernadero}

\begin{figure}[htbp]
  \centering
  \begin{subfigure}{0.9\linewidth} \centering
    \includegraphics[width=8cm]{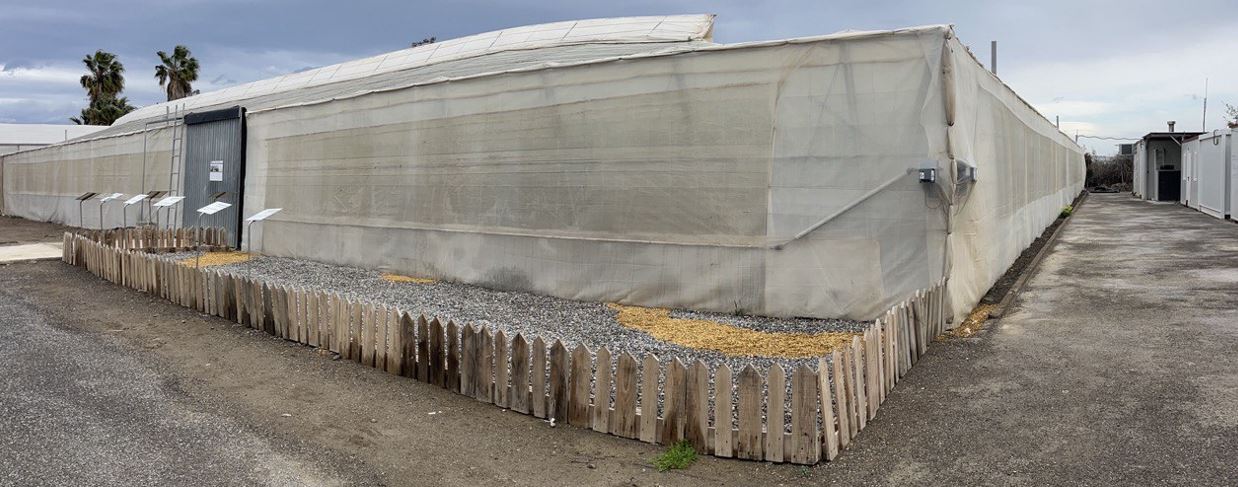} \centering
    \caption{Greenhouse outdoor}
    \label{fig:sub1}
  \end{subfigure}
  \begin{subfigure}{0.9\linewidth} \centering
    \includegraphics[width=8cm]{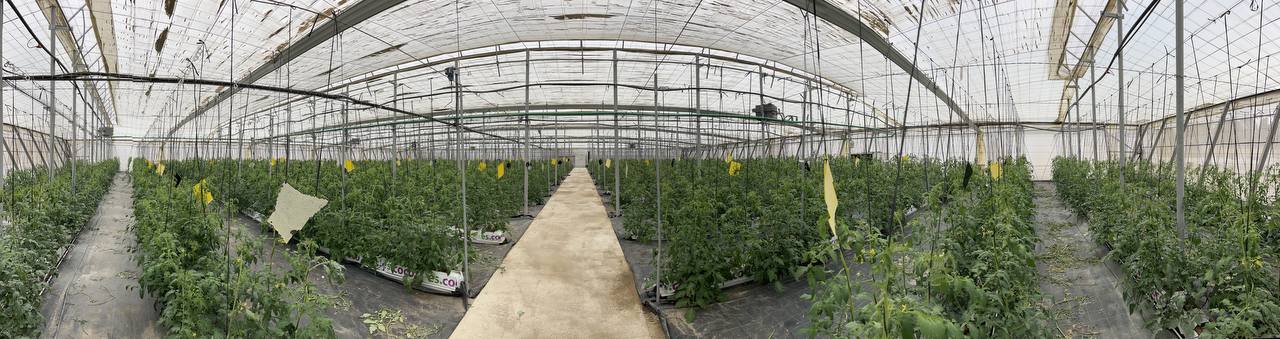} \centering
    \caption{Greenhouse indoor}
    \label{fig:sub2}
  \end{subfigure}
\caption{Agroconnect greenhouse} \label{fig:Fig1}
\end{figure}

The 3D model of the greenhouse used in this study was developed from the Agroconnect facilities in the municipality of La Cañada de San Urbano, Almería, Spain (Figure \ref{fig:Fig1}).

\begin{figure}[htbp]
\centering
\begin{subfigure}{0.8\linewidth} \centering
\includegraphics[width=7cm]{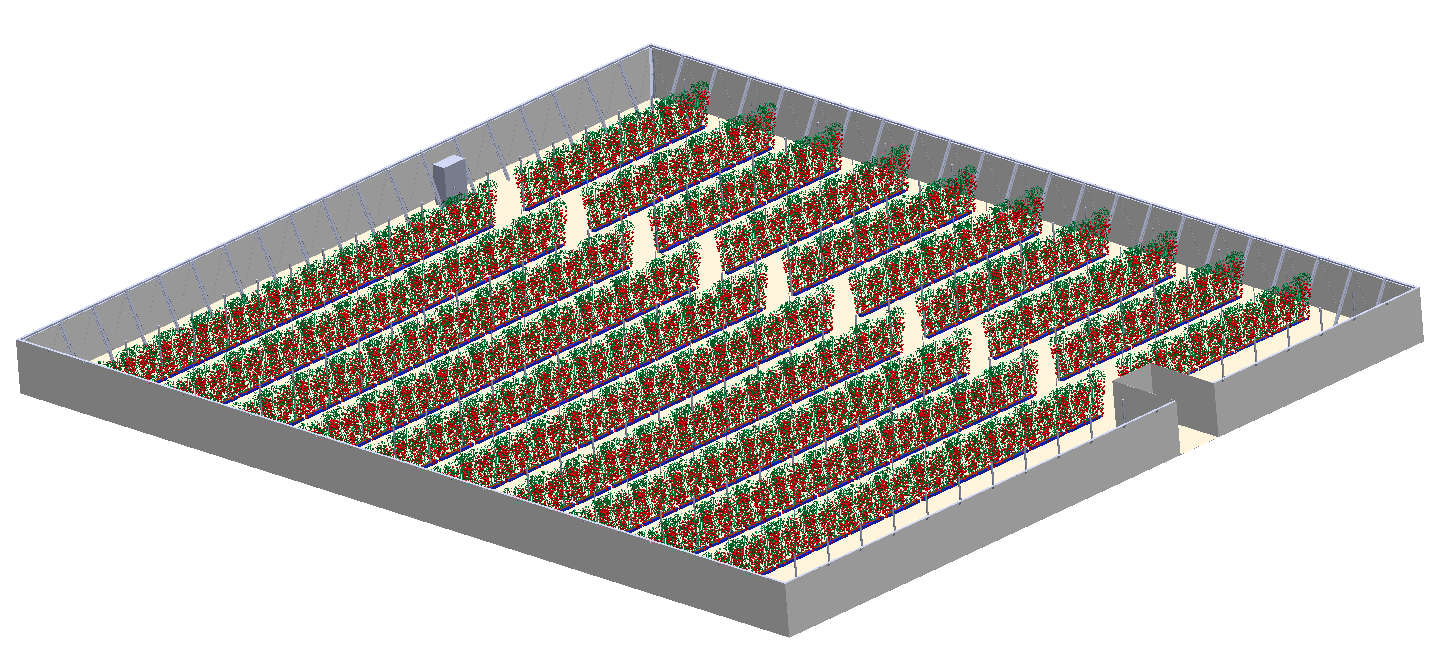} \centering
\caption{Agroconnect 3D greenhouse model}
\label{fig:Model3D}
\end{subfigure}
\begin{subfigure}{0.8\linewidth} \centering
\includegraphics[width=4cm]{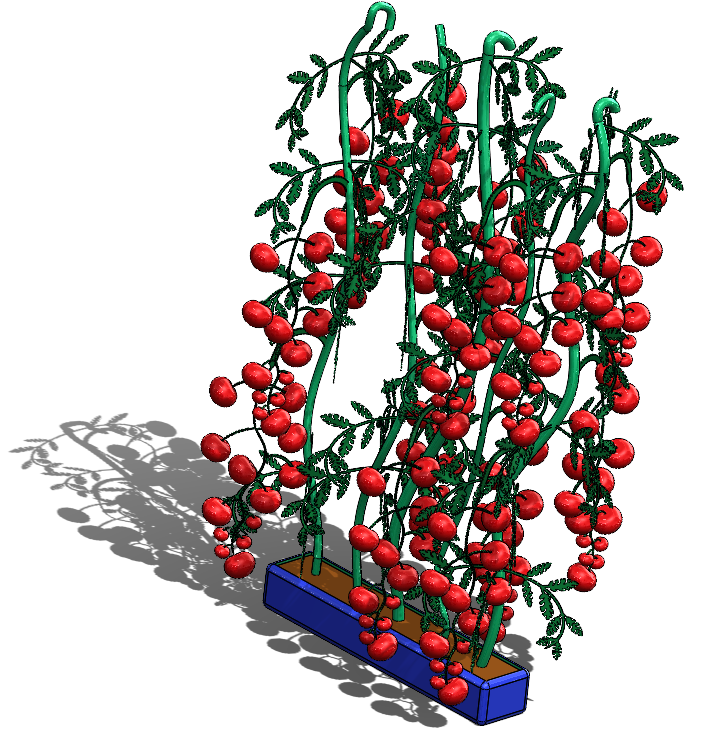} \centering
\caption{3D model of a tomato plant}
\label{fig:Tom}
\end{subfigure}
\caption{Agroconnect environment} \label{fig:Fig2}
\end{figure}

A 3D model of the greenhouse environment (Figure \ref{fig:Model3D}) was developed in SolidWorks, meticulously reproducing the real facility's architectural features. The model accurately captures the structural nuances of the greenhouse, including the primary support columns and diagonal reinforcement elements, which are faithfully represented in their real-world configuration. Furthermore, a full-scale 3D model of the tomato plant grown in the experimental facilities (Figure \ref{fig:Tom}) was incorporated. Both tomato plants and vertical and oblique support pillars were positioned to match their actual locations within the greenhouse.

For this work, a 20 × 20 m virtual greenhouse model was constructed, consisting of five corridors, each bordered on both sides by tomato plants and separated by 4 m, thereby defining the robot’s navigation lanes. The 3D model closely replicates a real pear-tomato crop grown in a hydroponic system and the characteristic geometry of a Mediterranean greenhouse. This realistic representation provides a reliable testbed for validating navigation algorithms previously applied in related research \citep{aranega2024navegacion,canadas2024multimodal,canadas2024autonomous}.

\subsubsection{Robot model} \label{3.1.2. Robots}

\begin{figure}[htbp]
    \centering
    \includegraphics[width=8cm]{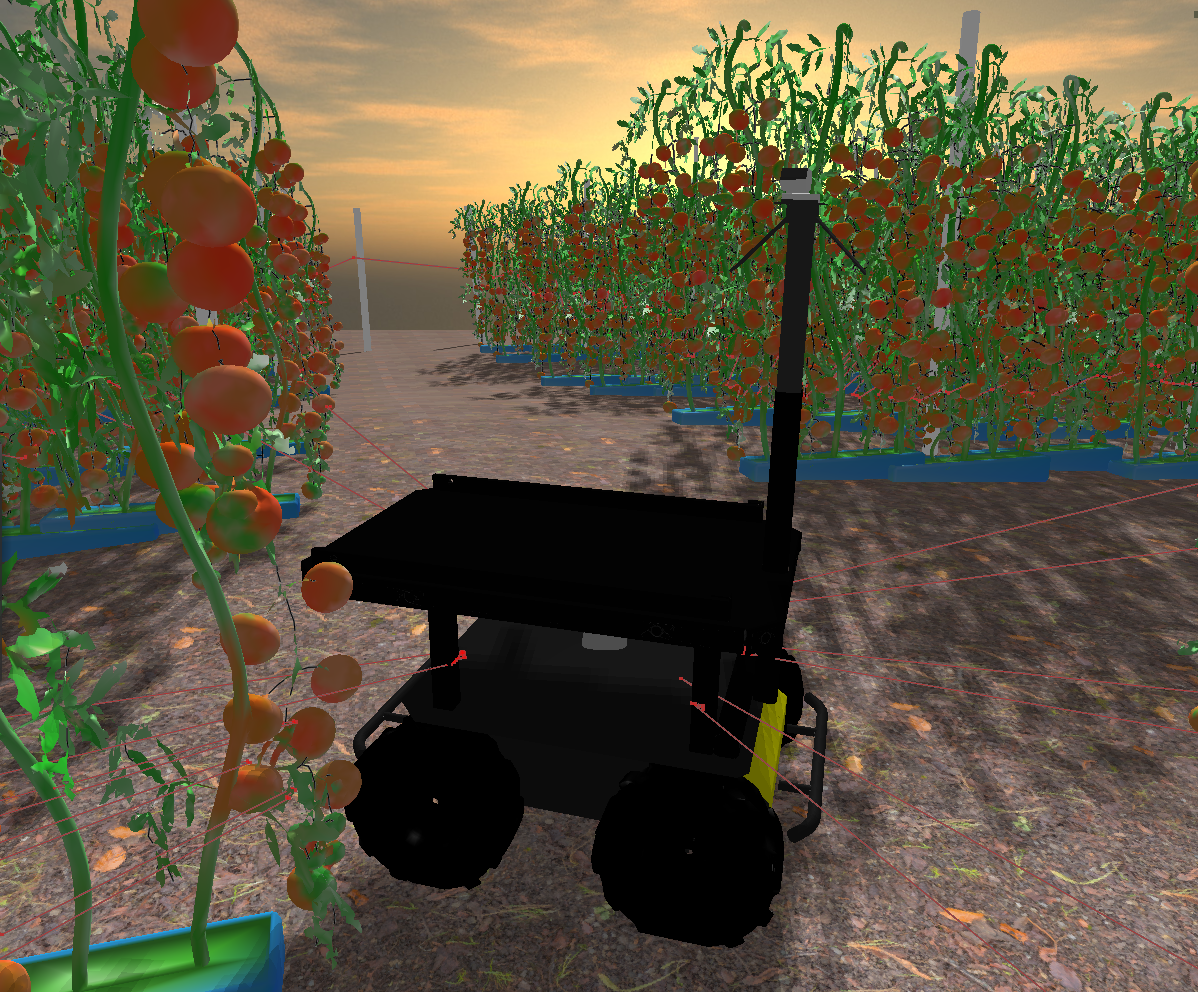}
    \caption{Differential robot ``Agricobiot I'' in MVSim}
    \label{fig:Smodel1}
\end{figure}

The benchmark employs a four-wheel differential robot configuration based on the Husky platform from Clearpath Robotics, although with several modifications specifically designed to operate inside a greenhouse and to facilitate interaction with the environment and agricultural workers. This modified platform was first presented in \cite{moreno2022modelado}, and its robustness and suitability for greenhouse applications have been demonstrated in several subsequent works (e.g., \cite{aranega2025nbv}). The robot has a total mass of 75 kg and is equipped with a model of 2D Hokuyo LiDAR \footnote{2D Hokuyo LiDAR: \url{https://www.hokuyo-aut.jp/search/single.php?serial=162}} sensor for high-level autonomous navigation, along with an onboard standard stereo camera used for environmental perception. The robot uses four wheels with V-shaped lugs, which provide optimal traction on sandy soil and prevent digging effects when ascending steep terrain. The real platform on which this model is based uses two 24 V DC motors with integrated encoders and a maximum power of 140 W, enabling the system to reach speeds of up to 1 m/s. Figure~\ref{fig:Smodel1} shows the robot model used in the simulation, including its main structural components and sensing devices.

This system constitutes the leading control platform, whose dynamics are described by the states of a differential robot $\mathbf{x}(t) = [\,x(t)\; y(t)\;\theta(t)\,] \in \mathbb{R}^{3}$, for $t \in \mathbb{R}^{+}$, as described as follows:
\begin{equation}
\begin{aligned}
\dot{x}(t) &= v(t)\cos(\theta(t)), \\
\dot{y}(t) &= v(t)\sin(\theta(t)), \\
\dot{\theta}(t) &= \omega(t),
\end{aligned}
\label{eq:diff_kinematics}
\end{equation}

\noindent where $v(t) \in \mathbb{R}$ denotes the input longitudinal linear velocity of the robot and $\omega(t)\in \mathbb{R}$ is the input angular velocity of the robot. For low-level control, the simulator reproduces the dynamics of the engine's viscous friction $b\in \mathbb{R}$ in $\mathrm{N\, m \, rad^{-1} \, s^{-1}}$, and wheel inertia $J \in \mathbb{R}$ in kg$\,$m$^2$, taking into account the mass of the chassis of the robot $m_{v}\in \mathbb{R}^{+}$ and the wheels $m_w\in \mathbb{R}^{+}$, the sum being the total mass $m_{robot} \in \mathbb{R}^{+}$ in kg. The total $J$ combines the rotational energy of the wheels and vehicle $I_{yy}\in \mathbb{R}^{+}$, together with the effects of dynamic friction and damping $C_D\in \mathbb{R}^{+}$. The resulting dynamics are expressed in \eqref{eq: mvsim22}, where the torque applied by the motors $\tau_{m,i}\in \mathbb{R}$ in N$\,$m and their torque constant $k_\tau\in \mathbb{R}$ are related to the angular velocity of the drive wheels $\omega_i\in \mathbb{R}$ rad/s, with $i \in \{R, L\}$ where $R$ represents the right motor and $L$ represents the left motor.

\begin{equation}
J\,\dot{\omega}_i(t) + b\,\omega_i(t) = k_\tau\,\tau_{m,i}(t) .
\label{eq: mvsim22}
\end{equation} 

The value of $\omega_i$ is determined by the inverse kinematic equations, \eqref{eq:diff_control}, which convert the speed command from the trajectory control $v$ and $\omega$ into a reference speed $\omega_{ref,i}\in \mathbb{R}$ for each motor that takes into account the wheel radius $r\in \mathbb{R}^{+}$ and the separation length between them $L_w \in \mathbb{R}^+$.

\begin{equation}
\begin{aligned}
v(t) &= \frac{r}{2}\left(\omega_{ref,R}(t) + \omega_{ref,L}(t)\right), \\
\omega(t) &= \frac{r}{L_w}\left(\omega_{ref,R}(t) - \omega_{ref,L}(t)\right).
\label{eq:diff_control}
\end{aligned}
\end{equation}

In this platform, a single motor drives each pair of wheels on the same side through a tensioned pulley system. Consequently, $\omega_{ref,R}(t)$ represents the angular velocity of both right wheels and $\omega_{ref,L}(t)$ that of the two left wheels. Therefore, the final differential kinematic model used in the control law corresponds to that of a standard two-wheeled mobile robot. With this equation, the basis for the robot's dynamics model is established, and expanded using a friction model in the following section.

\subsubsection{MultiVehicle Simulator (MVSim)}

\begin{figure}[htbp]
\centering
\begin{subfigure}{0.8\linewidth} \centering
\includegraphics[width=7cm]{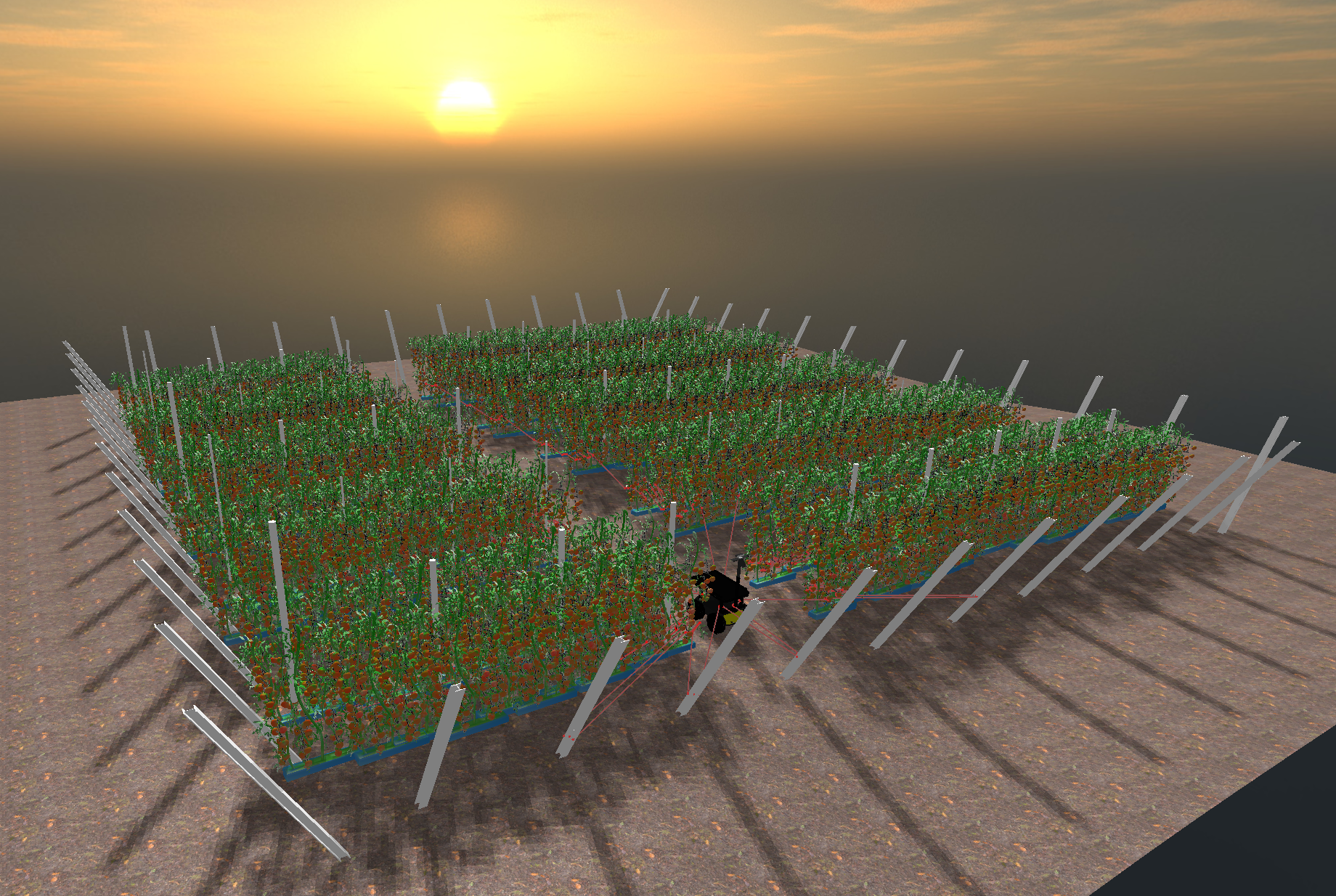} \centering
\caption{Complete greenhouse view}
\label{fig:Model3D1}
\end{subfigure}
\begin{subfigure}{0.8\linewidth} \centering
\includegraphics[width=7cm]{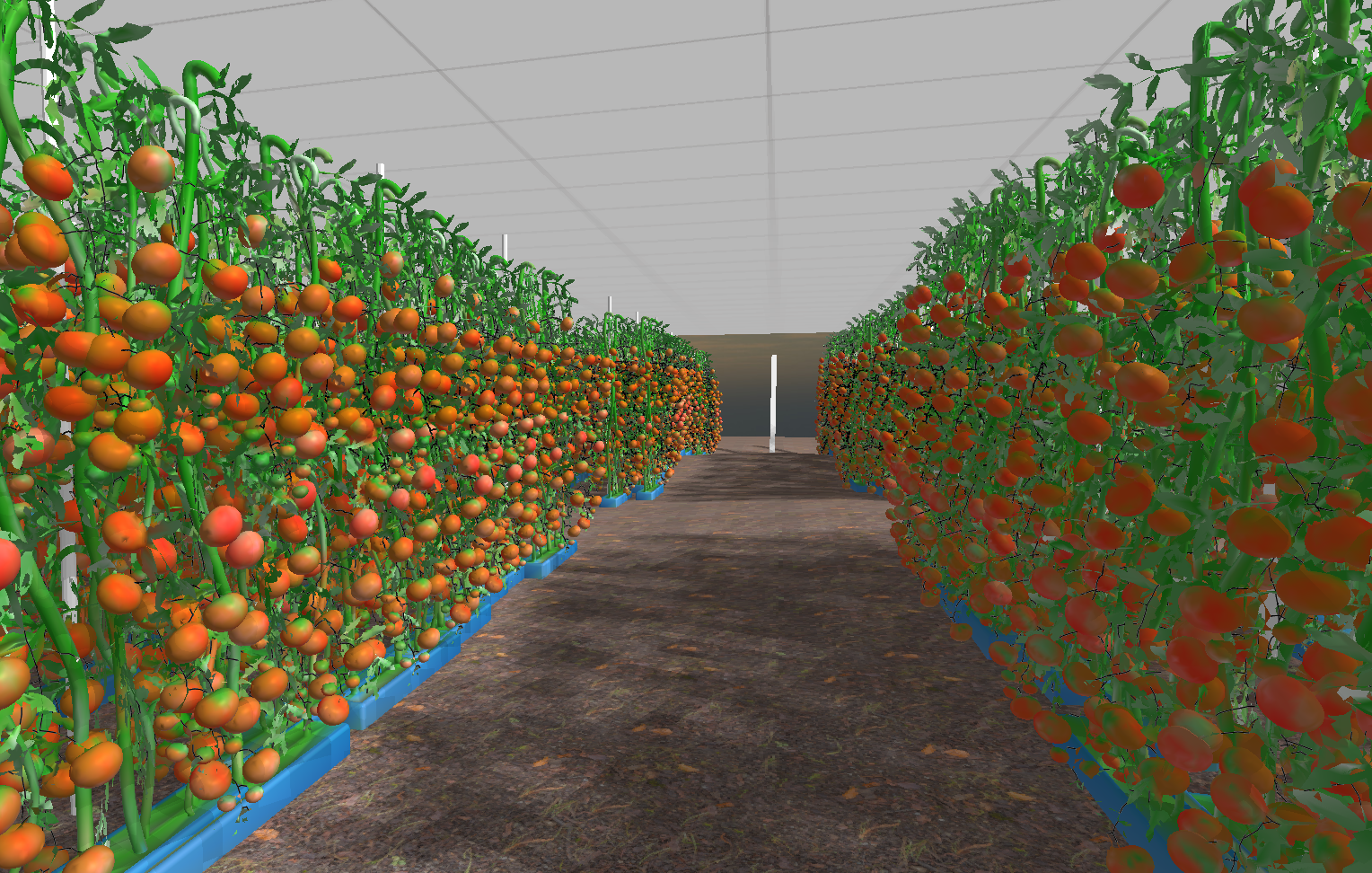} \centering
\caption{Corridor view}
\label{fig:Tom2}
\end{subfigure}
\caption{3D model in the simulator} \label{fig:Fig2b}
\end{figure}

The MultiVehicle Simulator (MVSim) \citep{blanco2023multivehicle} was used as the simulation environment, as it incorporates realistic physics-based friction models for the tyre-ground interaction, making it well-suited to analyse robot behaviour in this type of scenario. The greenhouse model created in SolidWorks was exported to Blender in Standard Triangle Language (\texttt{.stl}) format to apply color and visual enhancements, and subsequently exported to Digital Asset Exchange (\texttt{.dae}), a format widely used in robotics simulators. The resulting model is shown in Figure \ref{fig:Fig2b}, which presents both an external view of the greenhouse (Figure \ref{fig:Model3D1}) and the interior of the navigation corridors (Figure \ref{fig:Tom2}).

The MVSim friction model interprets the environment as an individual interaction between each wheel of the robot with the ground, decomposing the total forces $F_{t,i}(t) \in \mathbb{R}$  into longitudinal $F_{long,i}(t) \in \mathbb{R}^+$ and lateral $F_{lat,i}(t) \in \mathbb{R}^+$ components for each wheel. Before slippage occurs, the simulator evaluates the maximum admissible friction per wheel $F_{r_{max}}\in \mathbb{R}$ according to: 
\begin{equation}
F_{r_{max}} = \mu \, m_{pw}\, g,
\label{eq: mvsim3}
\end{equation}

\noindent where $\mu \in \mathbb{R}^{+}$ is the coefficient of friction between the wheel and the ground, $m_{pw} \in \mathbb{R}^{+}$ (obtained as $m_{pw}=m_{robot}/nW$, where $nW\in \{1, 2,3,4\}$ is the number of wheels, being 4 the maximum number of wheels allowed) is the effective mass per wheel in kg, and $g = 9.81\ \text{m/s}^2$ is the gravitational acceleration. Equation \eqref{eq: mvsim4} shows the lateral friction force,
\begin{equation}
F_{lat,i}(t) = a_{wy,i}(t)\, m_{pw},
\label{eq: mvsim4}
\end{equation}

\noindent where $a_{wy,i}(t) \in \mathbb{R}$ $\text{m/s}^2$ is the transverse acceleration of the wheel $i$. This force must be evaluated before exceeding $F_{r_{max}}$, establishing a limit so that there is no skidding as follows: 
\begin{equation}
F_{lat,i}(t) =
\operatorname{min}\big(F_{r_{max}},\, \max (-F_{r_{max}}, F_{lat,i}(t))\big),
\label{eq:F_rmax}
\end{equation}

Regarding longitudinal dynamics, \eqref{eq: mvsim5} calculates the longitudinal friction 

\begin{equation}
F_{long,i}(t) = \frac{1}{r} \, (\tau_{m,i}(t) - I_{yy}\, \omega_{{ref,i}}(t) - C_D \, \omega_i(t)),
\label{eq: mvsim5}
\end{equation}

\noindent which value is opposes the direction of travel of the robot, is obtained considering the torque developed by the motor $\tau_{m,i}(t)$, the inertia $I_{yy}$, the reference angular velocity $\omega_{ref,i}(t)$ rad/s and the damping component $C_D$. This value is also evaluated before exceeding the friction index $F_{r_{max}}$, preventing skidding and sliding when braking. The reduction in speed is represented by 
\begin{equation}
\alpha_i(t)=\frac{\tau_{m,i}(t) - rF_{long,i}(t) - C_D \omega_i(t)}{I_{yy}},
\label{eq: mvsim6}
\end{equation}
\noindent where $\alpha(t) \in \mathbb{R}$ is given in rad/s$^2$. Finally, the angular velocity is updated numerically as ($\Delta t$ is the simulation time step in s).
\begin{equation}
\omega_i(t)=\omega_i(t) + \alpha_i(t)\cdot\Delta t.
\label{eq: mvsim7}
\end{equation}

This is the equation that determines how the wheels' speed changes as a function of friction forces.

\subsection{Disturbances scenario}\label{sec: 3.2. setup}

In this section, the analysis focuses on a greenhouse, as it constitutes a particularly complex environment that poses several challenges for mobile robot control (in minimal room for manoeuvre and, consequently, a reduced margin for error, particularly in the presence of dynamic obstacles such as pedestrians, trolleys, or other moving agents) \citep{sanchez2024robotics}. In addition, Mediterranean greenhouses, particularly those in the Mediterranean region, commonly exhibit significant terrain variability, including irregular surfaces and changes in ground slope \citep{gonzalez2009navigation}. Such conditions directly influence all levels of the control architecture and may lead to navigation failures if not adequately addressed. To account for these factors, a set of representative disturbance scenarios is proposed, each introducing specific perturbations that users must consider when designing robust control strategies. This work examines three general situations of interest: (i) the robot’s payload, an unmeasurable disturbance, (ii) the characteristics of the ground surface, a measurable disturbance, and (iii) the inclination of the terrain, as a measurable disturbance. The objective of proposing measurable and non-measurable disturbances is to enrich the benchmark by increasing the number of strategies that can be implemented.

\subsubsection{Scenario 1. Load sensitivity} \label{3.2.1 Schenario 1}

The system's sensitivity to payload variations is directly linked to the increase in the normal force due to the robot's augmented total mass \citep{mehta2025deep}. This increase alters tire–ground interaction by expanding the contact area through terrain deformation. Consequently, a reduction in the robot's angular velocity is observed, since the increased load simultaneously affects the normal force, the moment of inertia, and the friction-related forces. In the simulator, this parameter influences all aspects of the dynamics, from the maximum admissible friction force, \eqref{eq: mvsim3}, to the longitudinal and lateral forces, \eqref{eq: mvsim4} and \eqref{eq: mvsim6}. Thus, a higher payload leads to an increase in the moment of inertia $I_{yy}$ and the longitudinal friction $F_{long}$, producing a larger change in angular acceleration $\alpha_i(t)$ and, consequently, in the resulting velocity. Since industrial load sensors are very expensive, it is configured as a fixed parameter at the start of the simulation. Therefore, it is configured as a parameter at the beginning of the simulation, and its influence is considered unknown and unexpected. In this paper, this value is fixed throughout the simulation and can range from $m_{pl}=0$ to $m_{pl}=70$ kg, as this is the maximum actual load that the robot can carry. From a mathematical perspective, this value varies with the type of transport, so it will be treated as a variable $p \in \{0, 70\}$ that changes with the payload, and will be included in the problem formulation.

\subsubsection{Scenario 2. Type of terrain} \label{3.2.2 Schenario 2}

\begin{figure}[htbp]
    \centering
    \includegraphics[width=\linewidth]{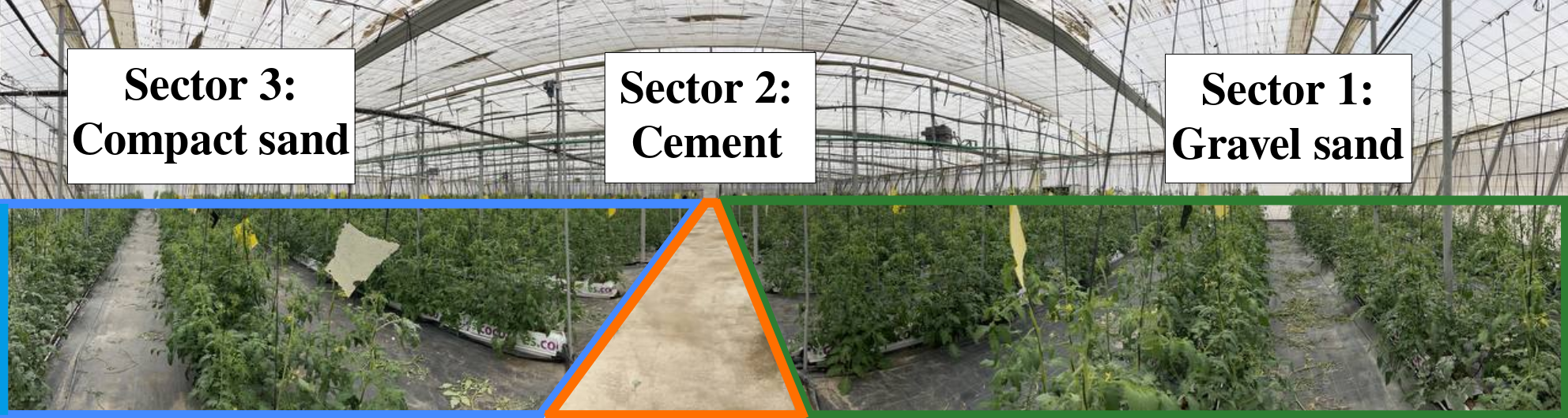}
    \caption{Soil type in the Agroconnect greenhouse}
    \label{fig:tterrain}
\end{figure}

Mediterranean greenhouses exhibit a wide variety of ground types that frequently include surface irregularities and slopes of up to 3--4\,\% \citep{canadas2024pid}. In adition, the type of soil found in a greenhouse also varies throughout the area, such as cement in areas where heavy machinery is often used, compacted sand where there is frequent foot traffic, or gravel where the soil has been tilled (see Figure \ref{fig:tterrain}). These conditions directly affect key physical parameters governing the robot’s dynamics, such as $\mu$, terrain slope $\phi \in \mathbb{R}$, $C_D$, and rolling resistance coefficient $C_{rr}\in \mathbb{R}^+$. This variability requires introducing an additional variable into the model equations, associated with the operating surface - cement, compact sand, or gravel - and denoted as $s$ ($s \in \{1,2,3\}$), which is updated in real time during the simulation depending on the type of terrain over which the robot is moving.

\begin{figure}[htbp]
\centering
\includegraphics[width=\linewidth]{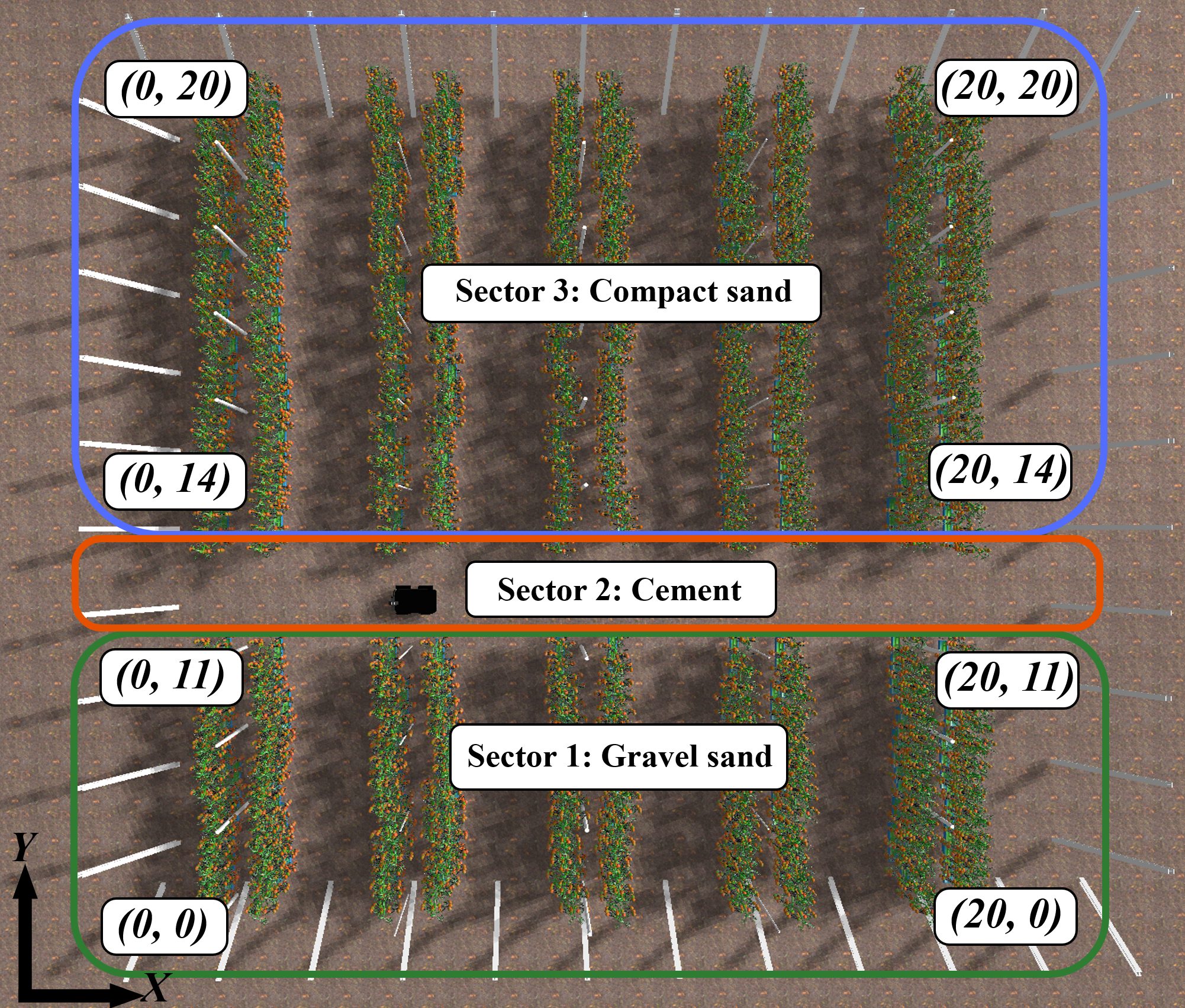}
\caption{Distribution of greenhouse terrain sectors (in meters).}
\label{fig:Fig5}
\end{figure}

As the values of these parameters change throughout the simulation, the soil-dependent variable $s$ is introduced and updated over time. This variable represents the influence of soil as a disturbance based on the area where the robot is located. For this reason, in order to replicate the actual conditions of the Agroconnect greenhouse, three types of terrain are defined and distributed by area in the simulation scenario (Figure~\ref{fig:Fig5}). Since the soil type does not change within a greenhouse environment, the areas corresponding to different soil types are defined. Accordingly, the values of $\mu$, $C_{rr}$, and $C_D$ are updated in real time, depending on the robot location, and this disturbance is therefore considered measurable.

Table~\ref{tab:Crr} summarizes the friction, rolling resistance and damping coefficients assigned to each terrain type.


\begin{table}[!ht]
\centering
\begin{tabular}{lcccc}
\hline
\textbf{Terrain sector} & \textbf{Soil} ($s$) & $\mu$ & $C_\mathrm{rr}$ & $C_D$ \\
\hline
Gravel sand & 1 & 0.2 & 0.10 & 1.25\\
Concrete & 2 & 0.8 & 0.01 & 0.75\\
Compact sand & 3 & 0.5 & 0.05 & 1.00\\
\hline
\end{tabular}
\caption{Typical friction and rolling resistance coefficients for different terrain types \citep{wong2022theory}.}
\label{tab:Crr}
\end{table}


\subsubsection{Scenario 3. Slope of the terrain} \label{3.2.3 Schenario 3}

In agro-industrial environments, terrain variability is common and directly affects the dynamics of mobile robots. Such irregularities modify the rotational velocity of each wheel, alter the robot’s heading, and introduce positional errors. This phenomenon becomes particularly critical in platforms used for transport tasks inside greenhouses, where uneven terrain is intrinsic to the operational environment. Furthermore, these slope disturbances, $\phi$, influence vehicle energy consumption, thereby reducing usable working hours.

\begin{figure}[htbp]
  \centering
  \includegraphics[trim = 0mm 0mm 0mm 0mm,clip,width=0.95\linewidth]{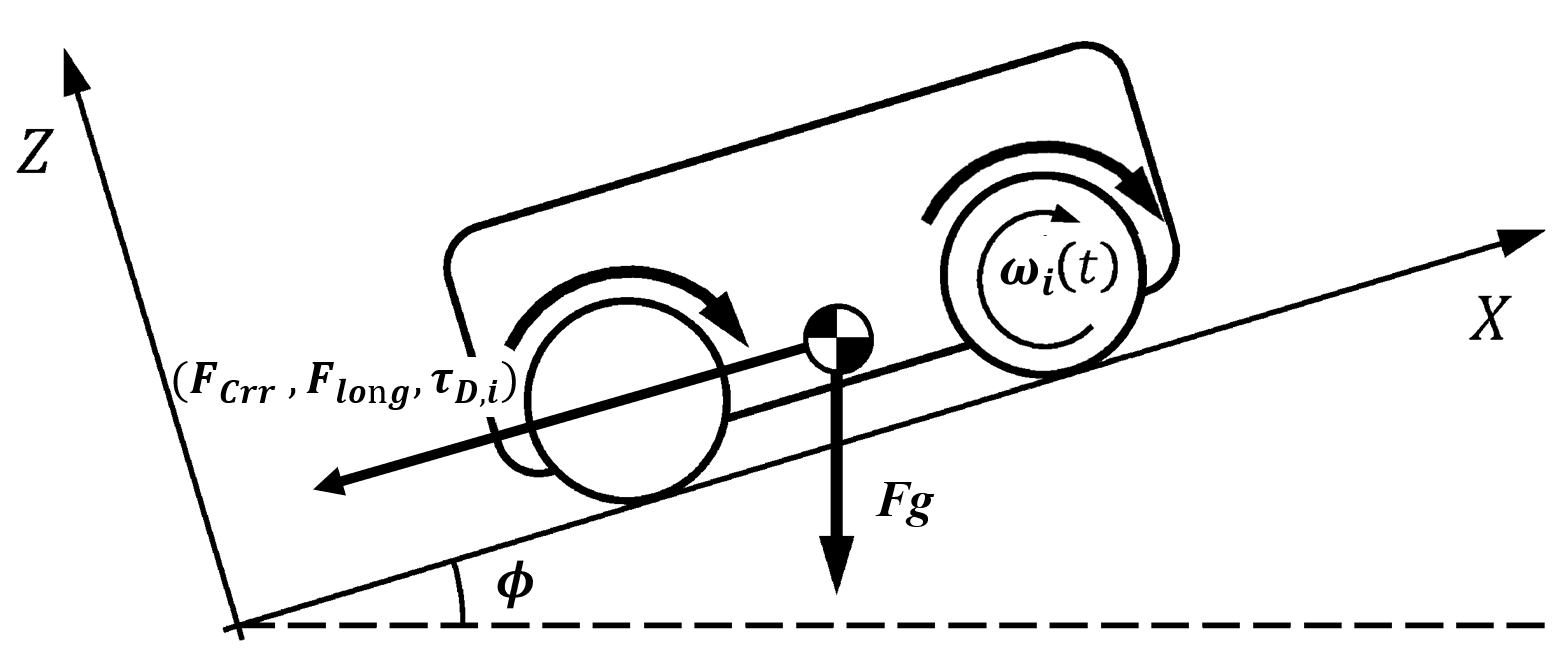}
  \caption{Characterization of greenhouse terrain.}
  \label{fig:inclinación}
\end{figure}

Therefore, a slope introduces a disturbance to the system as an additional resistive torque, whose magnitude depends on the angle of inclination and the robot's total mass (Figure \ref{fig:inclinación}). In a completely flat environment, the torque required to maintain a constant angular velocity is limited to overcoming friction and mechanical losses, which are often negligible. However, when a slope is present, the motor experiences a transient or sustained drop in angular velocity due to the additional load.

The slope angle $\phi(t)$ is estimated in real time using the \textit{pitch} value provided by the MVSim simulator, which is built upon the Mobile Robot Programming Toolkit (MRPT). This tool uses the \texttt{getPose3D} function, which computes the robot’s orientation $(\textit{roll}, \textit{pitch}, \textit{yaw})$ from simulated odometry, effectively emulating the behaviour of a real IMU equipped with magnetometers. Therefore, this disturbance is introduced into the simulator as a measurable quantity.

\section{Control Scheme Approach} \label{sec: 4. Low level}

This section describes the simulator's control scheme, where the user will understand and learn about the proposed law, divided into three layers: high-, mid- and low-level.

\subsection{Low-Level Control. PID} \label{4.1. Low level}

The low-level control focuses on the hardware layer within the robot’s control architecture.
Figure \ref{fig:lowlwvel} shows the final scheme implemented for this level, distinguishing between the robot's right and left motors. In this context, classical PID control, often combined with complementary techniques, remains one of the most widely adopted and robust solutions for DC motor regulation \citep{zhao2025pid}.\\

\textbf{\textit{PID controller}}\\

For the design of the internal loop, data from AgriCobIoT I (see Section 3.2) is obtained to calculate the robot's motor dynamics. The parameters $b$ and $k_\tau$  are used in \eqref{eq: mvsim22} to obtain the corresponding first-order model, \eqref{eq:lmodel}, with a static gain $K_m \in \mathbb{R}$ and a time constant $\tau_{motor} \in \mathbb{R}$. 
\begin{equation}
G(s) = \frac{K_m}{\tau_{motor} s+1}.
\label{eq:lmodel}
\end{equation}

Using this linear model, it is essential to ensure robustness against system variations and external disturbances. PID control is commonly applied to DC motors, defining an independent control loop for each of the robot’s driving wheels. In this case, \eqref{eq:tau_pid} presents the PID controller used to compute the motor torque $\tau_{PID,i}(t)\in \mathbb{R}$, based on the angular velocity error $e_{\omega,i}(t) \in \mathbb{R}$ as the difference between the reference speed $\omega_{ref,i}(t)$ and the current speed $\omega_i(t)$. The controller employs proportional, integral, and derivative gains ($K_P\in \mathbb{R}$, $K_I\in \mathbb{R}$, and $K_D\in \mathbb{R}$, respectively). The derivative term includes a low-pass filter to attenuate measurement noise, whose behavior is shaped by the filter parameter $N\in \mathbb{N}^+$ \citep{segovia2014measurement}.

\begin{equation}
\tau_{PID,i}(t) =
K_{p_i}\, e_{\omega,i}(t)
+ K_{i} \int_{0}^{t} e_{\omega,i}(\tau)\, d\tau
+ K_{d}\, \tau_{d,i}(t)
\label{eq:tau_pid}
\end{equation}
\\

\noindent where $\tau_{d,i}(t) \in \mathbb{R}$ satisfies the differential equation:

\begin{equation}
\dot\tau_{d,i}(t) +N\tau_{d,i}(t)=N\dot e_{\omega,i}(t)
\label{eq:tau_pid2}
\end{equation}

 
\textbf{\textit{Antiwindup technique}}\\

The motors exhibit a maximum effective torque of $\pm \tau_{max}\in \mathbb{R}$\,Nm, which defines the upper and lower saturation bounds of the inner control loop. To prevent integrator windup, an anti-windup mechanism is implemented as shown in equation \eqref{eq:antiwindup}. This mechanism introduces a feedback term that limits or corrects the integral action when saturation occurs between the action resulting from the PID controller and the feedforward mechanism (explained in the following point) $\tau_{mff,i} \in \mathbb{R}$ with the same action after saturation $\tau_{{m,i}_{sat}}=\min\big(\tau_{max},\, \max (\tau_{mff,i},- \tau_{max}\big)\big)\in \mathbb{R}$, avoiding excessive accumulation and unwanted dynamic responses \citep{aastrom2009control}. When the control signal saturates, the difference between the commanded torque and its saturated value is scaled by the tracking constant $K_{aw} \in \mathbb{R}$ as $ K_P / K_I$ and added to the integral term derivative $\dot{I}(t)\in \mathbb{R}$, effectively preventing windup.
\begin{equation}
\dot{I}(t) = K_Ie_{\omega,i}(t) + \frac{1}{K_{aw}} \big( \tau_{mff,i}(t) - \tau_{{m,i}_{sat}}(t) \big)
\label{eq:antiwindup}
\end{equation}\\
This is the equation that prevents this phenomenon, adding up at the moment before the integral (see Figure \ref{fig:lowlwvel}).\\

\textbf{\textit{Reference filter}}\\

Depending on the tuning method, closed-loop transfer functions may present undesirable zeros originating from the integral and derivative contributions. To address this, a reference filter is added, with the option of using a second-order formulation to cancel such zeros or to test more advanced cancellation strategies \citep{ast+haggISA2006, ogata2010modern}. Equation \eqref{eq:filter} shows the differential equation describing the filtering process, where $\omega_{fil,i}(t)\in \mathbb{R}$ represents the filtered reference. Also, it is possible to determine different orders with the value $n_f \in \mathbb{R}^+$, obtaining filters of different orders. The behaviour is determined by the filter time constant $\tau_f\in \mathbb{R}^+$ s.
\begin{equation}
\left(\tau_f \dot \omega_{ref,i}(t) + 1\right)^{n_f} \, \omega_{ref,i}(t)
= \omega_{fil,i}(t).
\label{eq:filter}
\end{equation}

With this equation, it is possible to determine a filtered signal.\\

\textbf{\textit{Feedforward technique}}\\

This section presents how an online estimation of the force acting on the wheels due to terrain slope, soil type, and the robot's payload can be performed. This estimation can be used to design a feedforward (FF) compensator to counteract the effects of terrain variations on the motor dynamics and, consequently, on the controlled angular velocity of the robot in real time. As illustrated in Figure \ref{fig:inclinación}, the main forces acting on the system must be understood in order to implement these various techniques:

\begin{figure*}[!t]
    \centering
    \includegraphics[width=\linewidth]{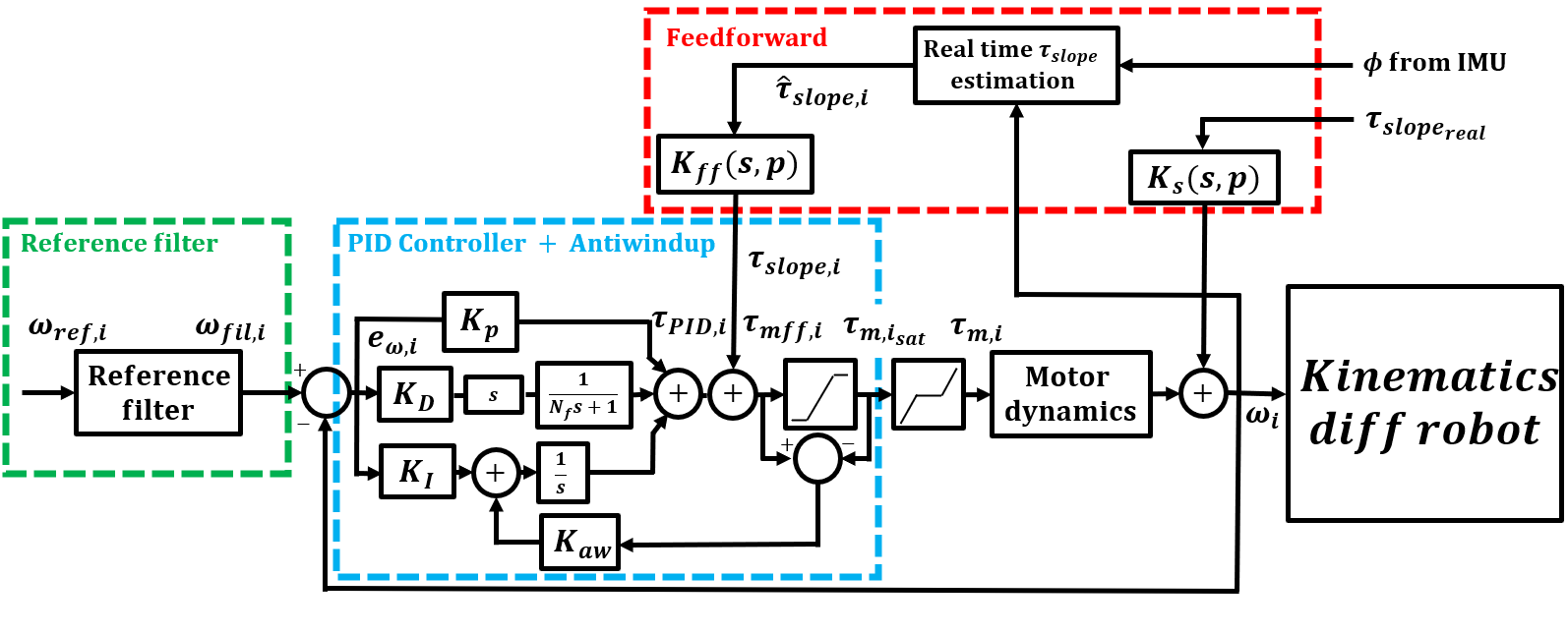}
    \caption{Low-level control scheme for motor $i \in \{R,L\}$.}
    \label{fig:lowlwvel}
\end{figure*}

\begin{itemize}
    \item Dynamical friction $F_{r_{\text{max}}}$: maximum admissible force before slipping occurs.
    \begin{equation}
    F_{r_{max}}(s,p)= \mu(s) \, m_{pw} (p) \, g .
    \label{eq:Fr}
    \end{equation}

    \item Rolling resistance $F_{C_{rr}}$: force opposing the motion due to tire rolling.
    \begin{equation}
    F_{C_{rr}}(s,p,\phi)= C_{rr}(s) \, m_{pw} (p) \, g \, \cos{(\phi)}\in \mathbb{R}.
    \label{eq:FCrr}
    \end{equation}

    \item Gravitational force $F_g$: tangential component of gravity associated with the terrain slope. 
    \begin{equation}
    F_g(p,\phi)=m_{pw} (p) \, g \, \sin{(\phi)}\in \mathbb{R}.
    \label{eq:Fg2}
    \end{equation}

    \item Slope force $F_{slope}$: As the total force of interaction between each wheel and the ground, evaluated with maximum friction (equation \ref{eq:Fr}). 
    \begin{equation}
    F_{slope}(s,p,\phi)=F_{C_{rr}}(s,p,\phi) + F_g(p,\phi) \in \mathbb{R}.
    \label{eq:Fg2b}
    \end{equation}

    \item Damping torque $\tau_{D}$: generated by axle damping and tire–ground interaction.
    \begin{equation}
    \tau_{D,i}(s,\omega_i)= C_D(s) \, \omega_i \in \mathbb{R}.
    \label{eq:Fg3}
    \end{equation}
\end{itemize}

\noindent where $s$ represents the soil type, and $p$ the robot payload, previously described. Finally, the value of $\tau_{slope}\in \mathbb{R}$ is estimated as shown in \eqref{eq:FT}, and this value is applied to each wheel.
\begin{equation}
    \tau_{slope,i}(s,p,\phi,\omega_i) = F_{slope}(s,p,\phi)\, r - \tau_{D,i}(s,\omega_i).
    \label{eq:FT}
\end{equation}

Once a real-time estimate of the force acting on the robot's wheels is obtained, several experiments can be conducted to quantify the impact of slope-induced disturbances on motor velocity. User can drive the system to a steady-state linear velocity, allowing observation of the relationship between the slope, the torque variation induced by the slope, and the resulting change in motor output. This process facilitates the identification of the static gain, denoted as \( K_{s}(s,p) \in \mathbb{R} \) and expressed in \( \mathrm{rad\, s^{-1} \, N^{-1}\, m^{-1}} \). This gain is defined as the variation in the process output, \( \Delta \omega_i \), relative to the disturbance's effect on the input torque, \( \Delta \tau_{slope_{real}} \in \mathbb{R} \), associated with the angle \( \phi \). By relating \( \tau_{slope} \) to \( \omega_i \) for calculating \( K_s(s,p) \), a more direct analysis of the slope's impact on the system input is achieved, resulting in more accurate values. Then, a static FF compensator for $K_s(s,p)$ can be calculated \citep{guzman2024feedforward}, with a adimensional static FF gain obtained as in \eqref{eq:feedforward_mech}. This value weights the estimated torque slope $\tau_{slope}(s,p,\phi,\omega_i)$ to act on the system input.
\begin{equation}
K_{ff}= - \frac{K_s}{K_m} \in \mathbb{R}.
\label{eq:feedforward_mech}
\end{equation}

\subsection{Mid-Level Control. Model Predictive Control} \label{3.2.1. Low level}

\begin{figure*}[!t]
    \centering
    \includegraphics[width=\linewidth]{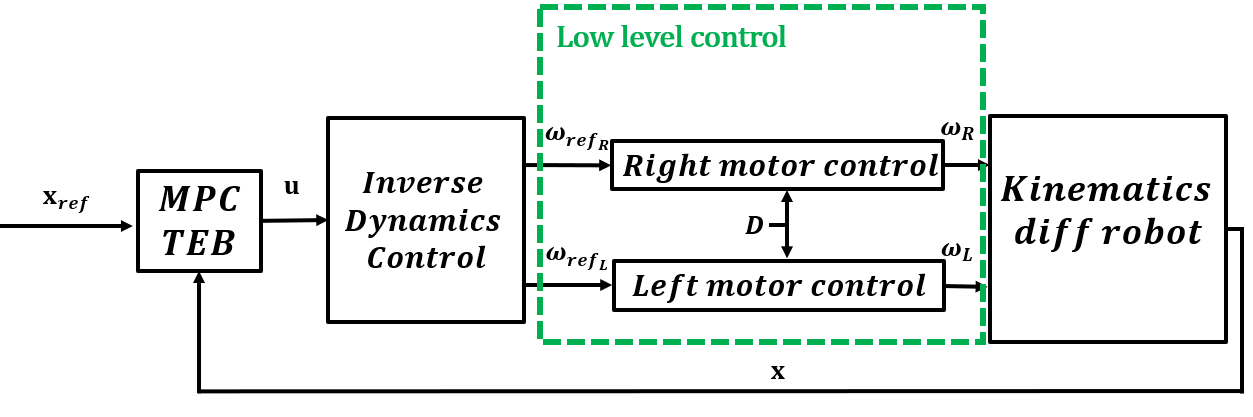}
    \caption{Mid-level control scheme. Low-level control is described in Figure \ref{fig:lowlwvel}.}
    \label{fig:mid_level}
\end{figure*}

The medium-level control focuses on trajectory tracking by employing an MPC strategy. Figure \ref{fig:mid_level} shows the proposed scheme implemented in this layer, where $\mathbf{D}$ corresponds to the value of the disturbances $(p,s,\phi)$. While the low-level controller is designed in the continuous-time domain using classical control techniques, the mid-level control strategy relies on a discrete-time formulation. This hierarchical structure allows combining fast local control with predictive optimization capabilities. 

The formulation presented below considers the behavior of a differential-drive mobile robot, whose motion is inherently discrete and nonlinear due to its dynamics between $\mathbf{x}$ and the input signal $u\in \mathbb{R}$ at every instant $k\in \mathbb{N}^+$
\begin{equation}
     \mathbf{x}_{k+1} = f(\mathbf{x}_k, \mathbf{u}_k),
    \label{eq:MPC1}
\end{equation}

It is assumed that the state is fully measurable and that the dynamic function $f$ governing the system evolution is known. The system is subject to constraints on both states and inputs, reflecting, for example, physical actuator limitations and bounds on the maximal admissible velocities. These state and input constraints are collected in the set $Z$, such that $(\mathbf{x}(k), \mathbf{u}(k)) \in Z$ for all $k$. The combination of admissible state--control pairs and obstacle-avoidance constraints with a safety margin $\delta_{\text{obst}} > 0 \in \mathbb{R}^+$ is then expressed as:
\begin{equation}
    Z_{\mathcal{O}}
    :=
    \{ (\mathbf{x}, u) \in Z \;\mid\; \mathrm{dist}(\mathcal{H}(x), O_j) \ge \delta_{\text{obst}},\; j \in \mathbb{N}_{1:n_o} \}.
    \label{eq:MPC5}
\end{equation}
\noindent Here, $\mathcal{H}(\mathbf{x}) \subseteq \mathbb{Y}$ denotes a polytopic over-approximation of the robot’s footprint within the workspace $\mathbb{Y}$, and $O_j \subseteq \mathbb{Y}$ represent the polygonal obstacles in the environment. \eqref{eq:MPC5} specifies that the pair $(\mathbf{x}(t), \mathbf{u}(t))$ must satisfy the admissible input constraints of the state encoded in $Z$, while simultaneously ensuring that the robot maintains a minimum safety distance from every obstacle $j \in \mathbb{N}_{1:n_o}$ through the condition $\mathrm{dist}(\mathcal{H}(x), O_j) \ge \delta_{\text{obst}}$, where $n_o\in \mathbb{R^+}$ is the number of obstacles.

Building upon this general MPC formulation, the proposed approach incorporates a Timed Elastic Band (TEB) structure to enhance real-time feasibility and adaptability. TEB introduces an elastic temporal parametrization for both the predicted trajectory and the control inputs \citep{rosmann2017integrated}. This allows the controller to dynamically modify the temporal spacing between consecutive predicted states, thereby achieving quasi-optimal behavior while preserving the sparse structure of the underlying optimization problem. Moreover, the TEB formulation naturally accommodates the kinematic and nonholonomic constraints characteristic of differential-drive robots \citep{rosmann2015timed}. Given a measured state $\mathbf{x}(t)$ and the admissible set $Z_{\mathcal{O}}$ defined in~\eqref{eq:MPC5}, the resulting MPC-TEB optimization problem penalizes state deviations, control effort, and elastic temporal variation. The cost function is defined over a prediction horizon $P \in \mathbb{N}$ as
\begin{equation}
\min_{\{\mathbf{x}_k\},\, \{\mathbf{u}_k\},\, \{\Delta t_k\}}
\sum_{k=0}^{P-1}
\ell(\mathbf{x}_k, \mathbf{u}_k, \mathbf{x}_{ref})
+ \lambda_T (\Delta t_k)^2,
\label{eq:MPC_TEB_cost}
\end{equation}

\noindent where the stage cost $\ell$ is positive definite and given by

\begin{equation}
\ell(\mathbf{x}_k,\mathbf{u}_k,\mathbf{x}_{ref}) 
= \|\mathbf{x}_k - \mathbf{x}_{ref}\|_Q^2 + \|\mathbf{u}_k\|_R^2.
\end{equation}
\noindent where $\mathbf{x}_k$ and $\mathbf{u}_k = (v_k,\omega_k)$ denote the predicted state and input $k$ steps ahead, respectively, $\mathbf{x}_{ref}$ represents the reference state to be tracked, and $Q$ and $R$ are the weight matrices that weight each action in the optimisation. The elastic time interval $\Delta t_k$ introduces flexibility in the temporal execution of the trajectory and is penalized by the weight $\lambda_T\in \mathbb{R}$. The predicted states evolve according to the discretized nonlinear dynamics of the differential-drive robot, subject to:
\begin{equation}
\begin{aligned}
x_{k+1} &= x_k + \Delta t_k\, v_k \cos\theta_k,\\
y_{k+1} &= y_k + \Delta t_k\, v_k \sin\theta_k,\\
\theta_{k+1} &= \theta_k + \Delta t_k\, \omega_k,
\end{aligned}
\qquad k = 0,\dots,P-1,
\label{eq:dynamics}
\end{equation}
\noindent subject to the admissible state--input constraints
\begin{equation}
(\mathbf{x}_k, \mathbf{u}_k) \in Z_{\mathcal{O}}, \qquad k = 0,\dots,P-1,
\end{equation}
which enforce both actuator limits and obstacle-avoidance constraints. For completeness, velocity bounds are imposed as
\begin{equation}
|v_k| \le v_{\max}, 
\qquad 
|\omega_k| \le \omega_{\max},
\label{eq:velocity_limits}
\end{equation}
and the robot maintains a minimum safety $d_{\mathrm{safe}} \in \mathbb{R}^+$ together with the effective radius of the robot's footprint $r_{\mathrm{robot}} \in \mathbb{R}^+$ and the effective radius of the object $r_{O_j} \in \mathbb{R}^+$ must be less than the difference between the position of the robot $\mathbf{x}$ and that of each obstacle $\mathbf{p_{o_{O_j}}} \in \mathbb{R}^3$ 
\begin{equation}
\|\mathbf{x}_k - \mathbf{p_{o_{O_j}}}\|
\ge
r_{\mathrm{robot}} + r_{O_j} + d_{\mathrm{safe}},
\qquad j \in \mathbb{N}_{1:n_o}.
\label{eq:collision}
\end{equation}

This yields a unified formulation in which the stabilising behaviour of MPC is combined with the elastic temporal parameterisation of TEB, enabling robust and flexible trajectory tracking under dynamic and constrained environments. The magnitude $\delta_{\text{obst}}$ is determined directly from the Hokuyo 2D LiDAR sensor model used as discussed above. This sensor generates a two-dimensional map of the environment which, using the \texttt{costmap\_converter} package, is transformed into a polygonal representation of the obstacles. Subsequently, the \texttt{local\_costmap} used by TEB provides an explicit identification of these obstacles and, consequently, a direct estimate of the minimum distance between the robot and each of them.

\subsection{High-Level Control. Lazy \texorpdfstring{$\theta^*$}{theta*}} \label{3.2.3. High level}

At the higher level of the navigation architecture, global path planning is performed to provide the local TEB-MPC controller with a feasible geometric reference. Path planning in continuous three-dimensional (3D) environments is often more challenging than in continuous two-dimensional (2D) settings \citep{huang2024joint}. In 2D scenarios with polygonal obstacles, the shortest feasible paths can often be obtained by applying $A^*$ search over visibility graphs \citep{lozano1979algorithm}. Lazy Theta$^*$ is an efficient variant of the Theta$^*$ algorithm that computes shorter and smoother paths than those obtained with $A^*$, while maintaining a reduced computational cost. Its main contribution lies in its deferred evaluation strategy: line-of-sight checks are performed only when a node is about to be expanded, rather than during node generation \citep{nash2010lazy}.

For global path computation, the environment is discretized into a grid, yielding a set of $n_{max}\in \mathbb{N}^+$ nodes at the intersections of grid cells. Each node $n\in \mathbb{N}^+$ stores a cost value $g(n)$ representing the minimum known distance from the initial node to $n$. The heuristic function used to guide the search is defined in~\eqref{eq:theta1}, where $h(n)$ denotes the Euclidean distance to the goal node $n_{\mathrm{goal}}\in \mathbb{N}^+$:
\begin{equation}
u_{ref}(n) = g(n) + h(n) \in \mathbb{R}.
\label{eq:theta1}
\end{equation}

Lazy $\theta^*$ employs the function $\mathrm{LineOfSight}(n, n')$, which verifies whether the straight-line segment connecting these nodes intersects any obstacle. This criterion enables the algorithm to skip intermediate grid nodes, producing smoother global paths. Two cases arise:

\begin{itemize}
    \item \textbf{Direct line of sight:} If direct visibility exists between $\mathrm{parent}(n)$ and $n'$, the cost is updated as
    
    \begin{equation}
    \begin{aligned}    
    g(n') =& ~g(\mathrm{parent}(n)) + c(\mathrm{parent}(n), n'),
    \\[6pt]
    \mathrm{parent}(n') =& ~\mathrm{parent}(n). 
    \end{aligned}
    \end{equation}

    \item \textbf{No direct line of sight:} If no visibility exists, the update is performed following the grid structure:
    \begin{equation}
        g(n') = g(n) + c(n,n'),
        \qquad 
        \mathrm{parent}(n') = n.
    \end{equation}
\end{itemize}

The algorithm accepts a cost update only if it improves the previously computed value of $g(n')$, evaluated in the total number of corners $c_o\in \mathbb{R}^+$. Furthermore, this algorithm incorporates a weighting factor  $w_{euc}\in \mathbb{R}$, which scales the Euclidean distance cost used in the trajectory evaluation, and a weighting factor $w_{traversal}\in \mathbb{R}$, which penalizes node traversal based on their distance from the goal, discouraging unnecessarily long or inefficient paths during the optimization process. This procedure constitutes the mathematical core of Lazy Theta$^*$ and explains its ability to generate global paths that are both shorter and smoother than those produced by $A^*$, while preserving high computational efficiency. The planned trajectory is given by the ordered sequence

\begin{equation}
\mathbf{x}_{plan} = \left\{ \mathbf{x}(k) \right\}_{k=n_{start}}^{n_{goal}} \in (\mathbb{R}^3)^{m+1},
\end{equation}

where $m\in \mathbb{N}$ is the number of nodes on the path found in the proposed solution. This result is interpreted as the geometric reference for the TEB-MPC controller, which refines it into a dynamically feasible, time-parametrized trajectory that respects system dynamics and the constraints encoded in $Z_{\mathcal{O}}$. 

\begin{figure*}[htbp]
    \centering
    \includegraphics[width=\linewidth]{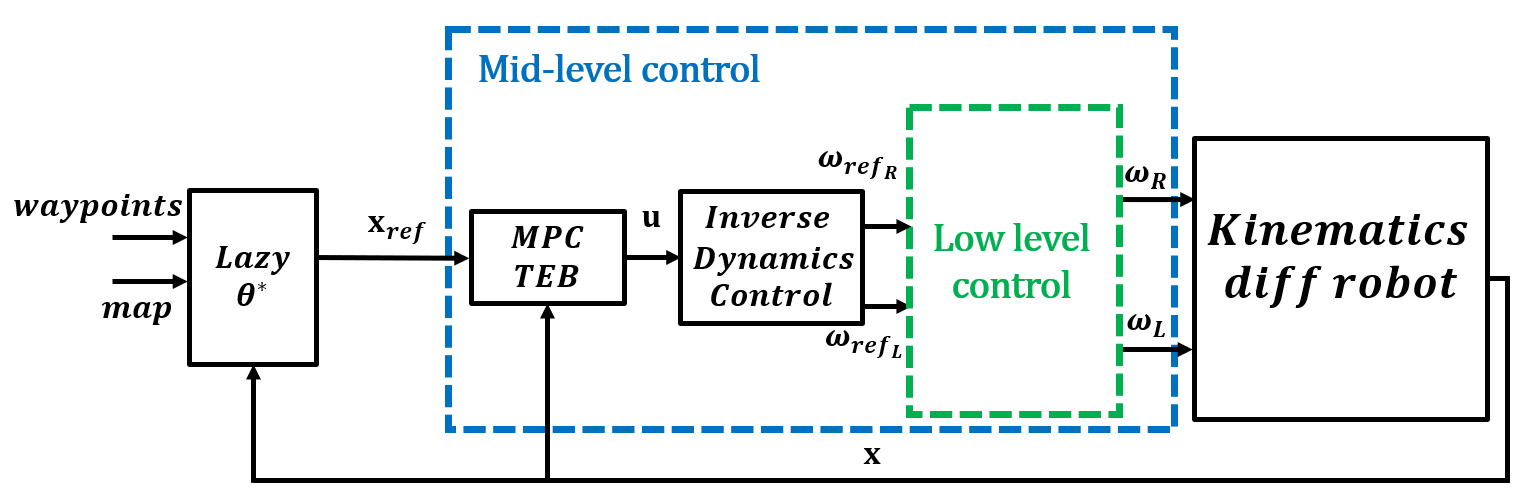}
    \caption{General high-level control scheme}
    \label{fig:hig_level}
\end{figure*}

Figure \ref{fig:hig_level} shows the final control diagram, combining the control schemes shown in Figures \ref{fig:lowlwvel} and \ref{fig:mid_level}, respectively.

\section{Benchmark challenge} \label{sec: 5. Competicion}

This section presents the tests prepared to evaluate the proposed controllers as part of the benchmark challenge covering the different scenarios described in Section \ref{sec: 3.2. setup}.

\subsection{Categories}

This section presents the experiments designed to evaluate the proposed controllers. The tests are organized into different categories, allowing a structured assessment of the scenarios introduced in the previous section. All of them report a \texttt{result/category\_x/yyyy\_mm\_dd\_hh\_mm\_ss.csv} file in the root directory where the variables of interest are stored, to provide post-processing if required.

\subsubsection{Category 1}

The first category focuses exclusively on evaluating the low-level controller, targeting users with a PID-based control level. The objective is to analyse the tracking accuracy of the wheel angular velocities ($\omega_{ref_L}$, $\omega_{ref_R}$). The total simulation time and the input reference $\mathbf{u}(t)$ applied to the low-level controller can be configured. Additionally, the user may adjust actuator-related parameters, including saturation limits and dead-zone effects, and enable or disable optional control features such as filtering, anti-windup compensation, and feedforward action. The corresponding tunable gains include $K_{P}$, $K_{I}$, $K_{D}$, $N$,$\tau_f$, $n_f$, $\tau_{max}$ and $K_{ff}$, which allow a detailed assessment of their impact on the closed-loop response. In addition, the user may: i) select the desired payload between 0 and 70 kg, ii) enable terrain change, iii) enable a slope between -4 and 4 $\degree$, combining them as desired. The user may conduct any experiments they wish with the proposed control law or may use their own control laws, a method that will be explained in the next section.

\subsubsection{Category 2}

The second category evaluates both the mid- and low-level control layers and is intended for users with expertise in process optimisation and predictive control. In this mode, the user can analyse the behaviour of the low-level motor controller and the implemented MPC simultaneously. All functionalities described in the first category remain available, while additional configuration options allow modifying the optimiser's cost weights $\lambda_T$, the obstacle-avoidance distance $d_{safe}\in \mathbb{R}^+$, weight matrices $Q$ / $R$, the success tolerance, the value of restrictions $(v_{max}, \omega_{max})$ and the sequence of waypoints the robot must reach. Furthermore, any combination of scenarios from Section \ref{sec: 3.2. setup} can be selected, and the simulation automatically adapts to these conditions. Users may evaluate the proposed MPC--TEB formulation or replace it with their own desired control strategy to compare performance under identical conditions.

\subsubsection{Category 3}

The third category evaluates the complete hierarchical control architecture, which includes the control layers at the high-, mid-, and low-level. It is designed for users with experience in hierarchical optimisation and global planning. In addition to all the configuration options available in the previous categories, this level allows modifying $n_{max}$ used to expand the graph, adjusting the cost associated with the trajectory length $w_{euc}$, and penalising high-cost $w_{traversal}$ nodes within the global planner. This category also supports any combination of scenarios. A two-dimensional coordinate vector is provided to the global planner, waypoints, together with a preloaded map required to evaluate the Lazy~$\theta^*$ algorithm. This example illustrates that the benchmark is suitable both for educational purposes and for advanced research. At this level, users may employ the provided global planner or substitute it with a custom implementation.

\subsection{Metrics} \label{5.2.1. Metrics}

To assess the quality of the implemented control law, the following performance indices are used:

\begin{itemize}
    \item Squared Absolute Error (SAE$_{nc}$): this metric represents the absolute value of the error over time, providing a direct measure of the controller’s mean tracking accuracy depending on the category number $nc \in \{1,2,3\}$.
    \begin{equation}
    \mathrm{SAE}_{nc} = \sum_{k=1}^{N} \left| e_{k,nc} \right|,
    \label{eq:SAE}
    \end{equation} 
    where \( e_{k,nc} \) is the error at time $k$ for \( nc \), and \( N \in \mathbb{N}^+ \) is the total number of samples.

    \item Squared Control Input (SCI$_{nc}$): this index quantifies the smoothness of the control signal by computing of the squared variations of the control input between consecutive samples:
    \begin{equation}
    \mathrm{SCI}_{nc} = \sum_{k=2}^{N} \left( u_{k,nc} - u_{k-1,nc} \right)^2,
    \label{eq:SCI}
    \end{equation}
    where \( u_{k,nc} \) denote as control input at time k for each of the categories \( nc \). Lower SCI values indicate smoother control actions and reduced actuator effort.
\end{itemize}

These performance indices are computed separately for each category of the benchmark, as detailed below.

\subsubsection{SAE and SCI - Category 1}

The cost function defined for this category, $J_1$, is focused on low-level control, dealing with tracking error and control effort for the speed of the robot's motors. First, the performance index $\mathrm{SAE}_1$ is computed from the tracking error $e_{\omega, i}$ in $\mathrm{rad/s}$, and on the other hand, the index $\mathrm{SCI}_1$ quantifies the variation of the control input based on the motor torque $\tau_{m,i}$, expressed in $\mathrm{N\,m}$. Both quantities are calculated for each wheel,  $i \in \{R,L\}$, and obtained from the ROS topic \texttt{benchmark\_params}. The corresponding signals for this category to compute the metrics are the following, normalised with respect to $\omega_{\max}$ and $\tau_{\max}$ respectively:
\begin{equation}
\begin{aligned}
e_{k,1} &=
\frac{
\left|e_{\omega, R}(k)\right|+\left|e_{\omega, L}(k)\right|
}{2\omega_{\max}} , \\
u_{k,1} &=
\frac{\left|\tau_{m,R}(k)\right|+\left|\tau_{m,L}(k)\right|}{2\tau_{\max}}.
\end{aligned}
\label{eq:category1_metrics}
\end{equation}

The global performance index, $J_{1}$, is obtained as the sum of the two metrics, $\mathrm{SAE}_1$ and $\mathrm{SCI}_1$:
\begin{equation}
J_1
=\mathrm{SAE}_1
+
\mathrm{SCI}_1.
\end{equation}

\subsubsection{SAE and SCI - Category 2}

The performance index for this category combines low- and mid-level control objectives. As the mid-level control objectives are focused on trajectory tracking, the error $e_{2}$ is computed as the Euclidean distance between the current robot position $(x,y)$ and the predicted position provided by the TEB planner $(x_{\mathrm{teb}},y_{\mathrm{teb}})\in \mathbb{R}^2$, obtained from the ROS topics \texttt{/odom} and \texttt{/teb\_odom}. The low-level control input is given by $u = (v,\omega)$, expressed in $(\mathrm{m/s},\mathrm{rad/s})$.

The instantaneous tracking error is normalised with respect to the maximum admissible linear and angular velocities, $v_{\max}$ and $\omega_{\max}$, and is defined as

\begin{equation}
\begin{aligned}
e_{k,2} &=
\frac{
\sqrt{(x(k)-x_{\mathrm{teb}}(k))^2 + (y(k)-y_{\mathrm{teb}}(k))^2}
}{v_{\max}\,\Delta t},
\\[6pt]
u_{k,2} &=
\sqrt{
\frac{v(k)^2}{v_{\max}^2}
+
\frac{\omega(k)^2}{\omega_{\max}^2}
}.
\end{aligned}
\label{eq:e2_categ2}
\end{equation}

Then, considering (\ref{eq:SAE}), (\ref{eq:SCI}), and  (\ref{eq:e2_categ2}) an index for the mid-level control is calculated as $J_{_2}=\mathrm{SAE}_2+\mathrm{SCI}_2$ to quantify the trajectory tracking error and the control effort for it. Finally, the total performance index for this category that combines $J_1$ and $J_2$ is given by the following expression:
\begin{equation}
J_{T_{2}}
=
\frac{1}{2}
\left( J_1+ J_2\right).
\end{equation}

\subsubsection{SAE - Category 3}

The low- and medium-level controllers discussed above generate explicit control actions-such as linear speeds, angular speeds, or motor torques — where the dynamic error and the control effort over time are combined in a performance index. However, the case of the planner is conceptually different. A planner does not act directly on the physical system, but rather generates an optimal geometric trajectory in state space, usually without explicit temporal information or an associated control signal. Therefore, there is no control signal and, consequently, it is not possible to define a ‘control effort’ in the classical sense. Therefore, the use of an SCI-type metric for the planner is conceptually incorrect and may lead to misinterpretations of system performance \citep{siciliano2008springer}.

In this work, the planner's performance is evaluated exclusively using a geometric index, which quantifies the spatial discrepancy between the trajectory generated by the planner and the trajectory actually followed by the robot:
\begin{equation}
\begin{aligned}
e_{k,3}(k) =& 
\left\|
\mathbf{x}_{\text{robot}}(k) - \mathbf{x}_{\text{plan}}(k)
\right\|
\end{aligned}
\label{eq:ek3}
\end{equation}

This index represents a direct measure of the geometric quality of the plan and should not be interpreted as a measure of the control effort. For conceptual consistency, a cost function $J_3=\mathrm{SAE_3}$ is defined for this objective, where $\mathrm{SAE_3}$ is calculated based on (\ref{eq:SAE}) and (\ref{eq:ek3}).

Then, the following performance index for this category is proposed as a dimensionless sum the three cost functions for each control layer, $J_1$, $J_2$, and $J_3$, resulting in:
\begin{equation}
J_{T_{3}}
=
\frac{1}{3}\big(J_1 + J_2 +
J_3
\big).
\end{equation}

\subsection{Reference case} \label{4.2.2. Results}

A reference case is proposed for each of the categories and scenarios defined. For this baseline configuration, the following simulation parameters are selected:\\

\textbf{\textit{Category 1 test - Low-level control}}\\

The inner-loop control of the AgriCobIoT~I robot is modelled using \eqref{eq: mvsim22}, substituting the actuator’s real parameters with a mass of $m_{robot}=70\,\mathrm{kg}$ as the nominal operating value of the robot. Considering $k_{\tau} = 1$ N\, m, it yields a time constant of $\tau_{motor} = 2.97$ s and a motor gain of $K_m = 1.33\, $Nm$\,$rad$^{-1}\,$s$^{-1}$, for the first-order model \eqref{eq:lmodel}, resulting in:
    \begin{equation}
        M(s)=\frac{1.33}{2.97 s+1}
    \end{equation}
    
A critically damped PI velocity controller is designed via pole placement at $-0.6$ and $-31.25$ s$^{-1}$, obtaining $K_P = 70$ N$\, $m$\, $s and $K_I = 40$ N$\, $m using the pole assignment tuning method \citep{aastrom2009control}. 
    \begin{equation}
        C(s)=70+\frac{40}{s}
    \end{equation}
Given the actuator torque saturation of $[-400,\,400]$ N$\,$m, an antiwindup back-calculation scheme is implemented with $K_{aw} = K_P/K_I$ \citep{aastrom2013computer}. A reference filter is also added to mitigate overshoot induced by the controller zero, with a value of $\tau_f = K_P/K_I$ s, and $n_f=1$ \citep{aastrom2009control}. 
    \begin{equation}
        F(s)=\frac{1}{(70/40s+1)}
    \end{equation}
    
Finally, $K_{ff} = 0$ in order to analyze the complete effect of the disturbances. Table \ref{tab:2} shows a summary of the parameters used in the case.

\begin{table}[h]
\centering
\caption{Low-level control params}
\begin{tabular}{l c l c}
\hline
\textbf{Params} & \textbf{Value used} & \textbf{Params} & \textbf{Value used} \\
\hline
$J$ & 2.22 kg$\, $ m$^{2}$ & $K_P$ & 70 N$\, $m$\, $s\\
$b$ & 0.75 N$\, $m$\, $s$\, $rad$^{-1}$ & $K_I$ & 40 N$\, $m \\
$k_{\tau}$ & 1 N$\, $m$\, $ & $K_{aw}$ & $K_p/K_I$ s$^{-1}$ \\
$K_m$ & 1.33 N$\, $m$\, $s$\, $rad$^{-1}$ & $\tau_f$ &  $K_p/K_I$ s\\ 
$\tau_{motor}$ & 2.97 s & $n_{f}$ & 1 \\ 
$\tau_{max}$ & $\pm$400 N\,m & $K_{ff}$ & 0.0 \\
\hline
\end{tabular}
\label{tab:2}
\end{table}

With these values, the test is defined as follows. The initial pose of the robot is set to $(x,y,\theta) = (10.1, 3.0, 0.78)$, and the robot moves for a total duration of 60~s. The experiment consists of a step change applied at $t = 4$~s, where both wheel angular velocities, $\omega_L$ and $\omega_R$, are set to 0.75~rad/s during 1 s. This step input is maintained until $t = 34$~s, when the robot stops and receives angular velocity commands $\omega_L = 0.75$~rad/s and $\omega_R = -0.75$~rad/s in order to perform a $180^{\circ}$ rotation. Subsequently, the robot advances for an additional 8~s with an angular velocity of 0.75~rad/s, and, after 50 s, the robot stops.

Finally, for all experimental trials, the figures report only the results obtained for a payload mass of $m_{pl}=70$, as this represents the most challenging operating condition. Nevertheless, the result tables for all categories include the performance data corresponding to both $m_{pl}=0$ kg y $m_{pl}=70$ kg. It should be noted that the proposed framework allows the user to select any desired payload mass, enabling flexible and comprehensive performance evaluation.

\begin{figure}[htbp]
\centering
\includegraphics[width=\linewidth]{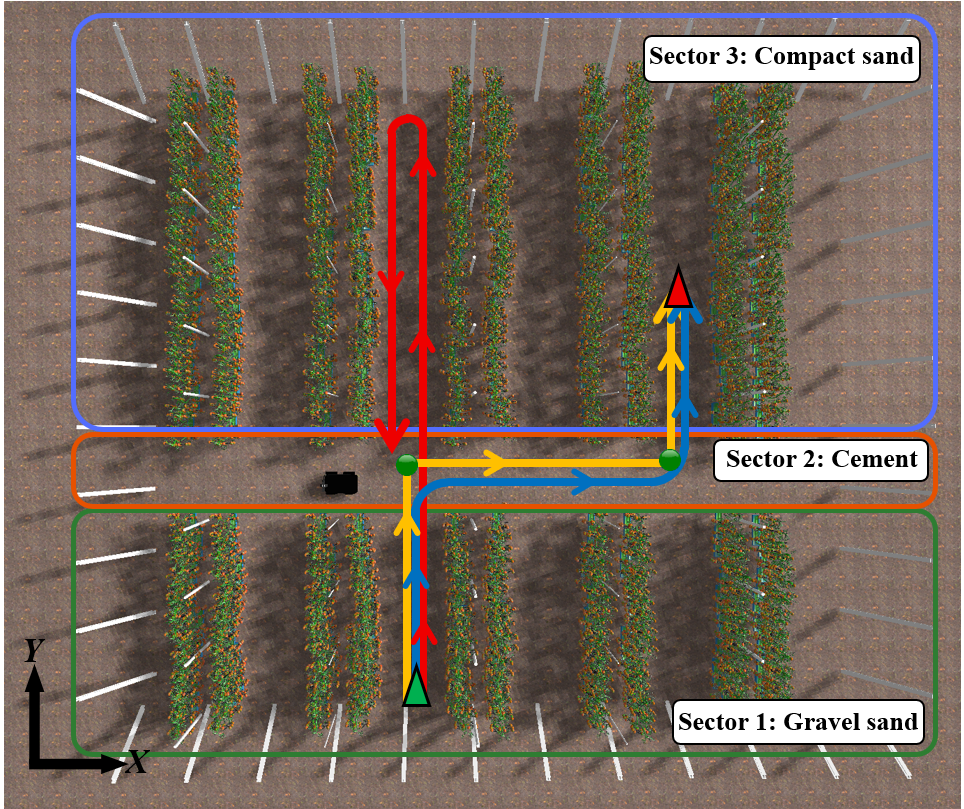}
\caption{A reference robot stroke is defined for each category, with a predefined initial orientation and a randomly generated final orientation. The green dot corresponds to the robot's initial pose, and points that follow category 2; the red arrow corresponds to the trajectory of category 1; the yellow arrow corresponds to the trajectory of category 2; the blue arrow corresponds to the trajectory of category 3; and the red cross corresponds to the target point for categories 2 and 3.}
\label{fig:Fig80}
\end{figure}

In this way, the robot traverses the three different terrain sectors associated with the maximum inclination profile. Figure~\ref{fig:Fig80} shows the stroke to follow by the robot, represented by a red arrow. Furthermore, this stroke is evaluated under two different payload conditions, namely 0 and 70~kg, considering four distinct scenarios: i) flat terrain without slope or terrain changes; ii) inclined terrain without terrain changes; iii) flat terrain with terrain changes; and iv) inclined terrain with terrain changes. It is important to note that, in the MVSim simulator, wheel encoders are modeled as ideal sensors by default; however, the user can introduce additive white Gaussian noise, allowing the encoder measurements to be expressed as the actual angular velocity corrupted by zero-mean Gaussian noise \citep{blanco2023multivehicle}. For this reason, three tests are carried out, yielding the mean and standard deviation for each.

Overall, this category provides a clear and representative benchmark for velocity control in a differential-drive robot, where only encoder feedback is required as the sensing modality. The encoder measurements are affected by a Gaussian noise model, which enriches the benchmark by introducing variability across runs and yielding non-identical results under repeated trials. The performance achieved in this category serves as a baseline for comparison with more advanced control strategies evaluated in subsequent categories.





\begin{figure*}[htbp]
\centering
\includegraphics[width=\linewidth]{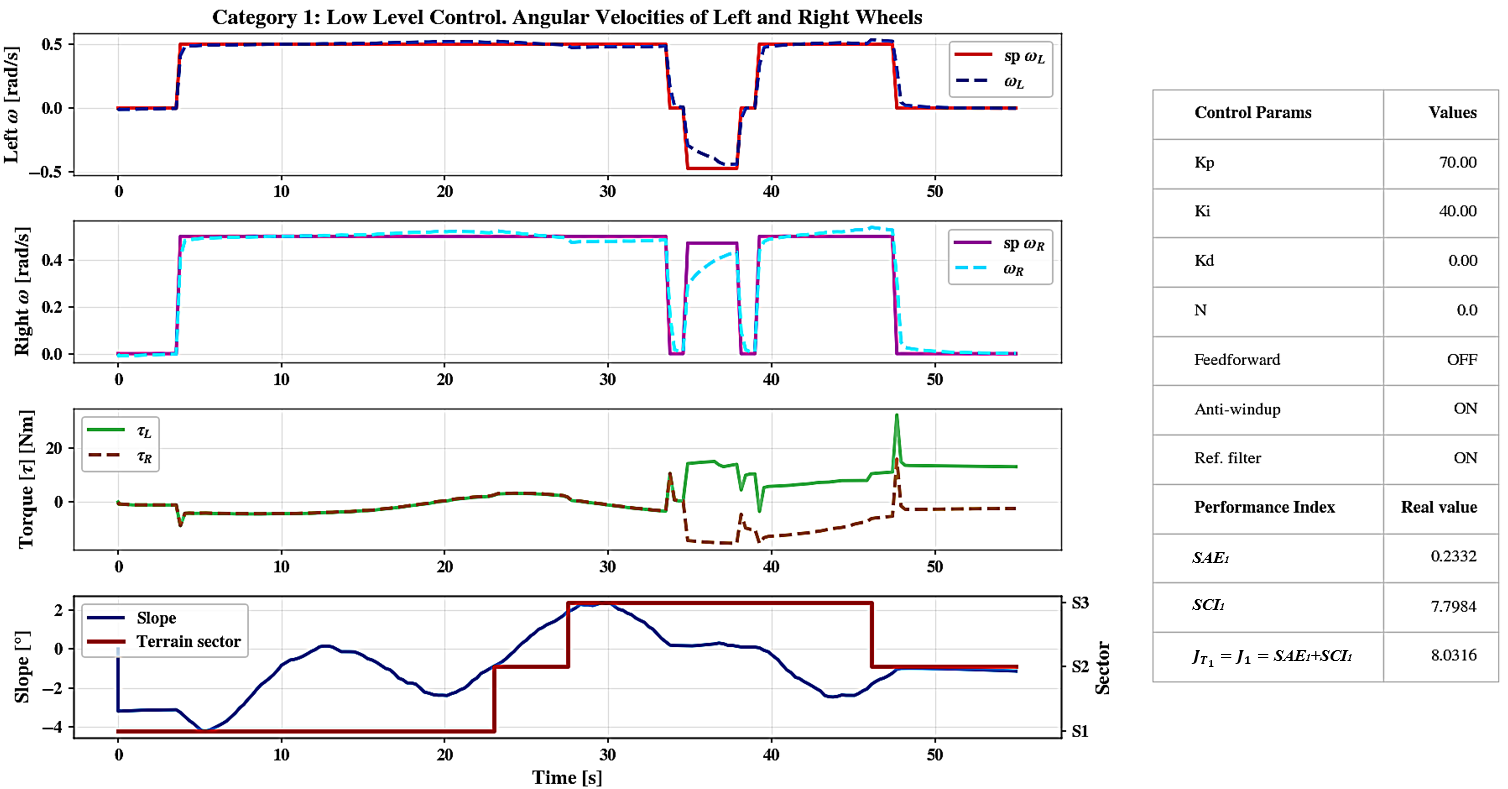}
\caption{Category 1 test, with $m_{pl}=70$, with slope on the ground and sector change}
\label{fig:c14}
\end{figure*}

\begin{table}[t]
\centering
\caption{Average SAE and SCI with standard deviation of 3 tests in Category~1 across all terrain and slope conditions.}
\label{tab:MASE_MSICJ}
\begin{adjustbox}{max width=\textwidth}
\begin{tabular}{ccccc}
\toprule
 &  & \multicolumn{3}{c}{\textbf{Low-level}} \\
\cmidrule(lr){3-5}
\textbf{Terrain Conditions} & \textbf{Payload [kg]} 
& $SAE_1$ & $SCI_1$ & $J_{1}$ \\ 
\midrule

\multirow{2}{*}{\shortstack{Slope OFF \\ Terrain OFF}} 
& 0  & 0.0988 $\pm$ 0.0048 & 7.9512 $\pm$ 0.3127 & 7.9602 $\pm$ 0.3381 \\
& 70 & 0.1192 $\pm$ 0.0061 & 8.0317 $\pm$ 0.3514 & 8.1509 $\pm$ 0.3726 \\ 
\midrule

\multirow{2}{*}{\shortstack{Slope ON \\ Terrain OFF}} 
& 0  & 0.1904 $\pm$ 0.0097 & 7.0156 $\pm$ 0.4012 & 7.2049 $\pm$ 0.4238 \\
& 70 & 0.2274 $\pm$ 0.0118 & 7.7686 $\pm$ 0.4885 & 7.9959 $\pm$ 0.5037 \\ 
\midrule

\multirow{2}{*}{\shortstack{Slope OFF \\ Terrain ON}} 
& 0  & 0.1008 $\pm$ 0.0043 & 6.0139 $\pm$ 0.3121 & 6.1240 $\pm$ 0.3347 \\
& 70 & 0.1106 $\pm$ 0.0057 & 6.6639 $\pm$ 0.3796 & 6.7745 $\pm$ 0.4012 \\ 
\midrule

\multirow{2}{*}{\shortstack{Slope ON \\ Terrain ON}} 
& 0  & 0.1849 $\pm$ 0.0091 & 7.5264 $\pm$ 0.4173 & 7.7159 $\pm$ 0.4418 \\
& 70 & 0.2332 $\pm$ 0.0127 & 7.7984 $\pm$ 0.4621 & 8.0316 $\pm$ 0.4875 \\ 

\bottomrule
\end{tabular}
\end{adjustbox}

\end{table}

The trajectories obtained for this category, together with the associated key performance indicators, are shown in Fig.~\ref{fig:c14}. Subplots 1 and 2 depict the setpoint and the angular velocities of the left and right wheels, respectively. The applied input torque is displayed in Subplot~3, while the terrain slope and the corresponding sector index are illustrated in Subplot~4. In addition to these real-time plots, a window displaying the MVSim simulator (as shown in Fig.~\ref{fig:Model3D1}) is presented during the simulation, allowing direct visual observation of the robot’s behavior.

The reported Key Performance Indicators (KPIs) are summarized in Table~\ref{tab:MASE_MSICJ}, characterizing the performance of the proposed strategy in terms of efficiency and operability. These indicators quantify the tracking error relative to the prescribed setpoint and the variability of the control signal, which is directly related to the robot’s energy consumption. Furthermore, the performance index provides insight into whether the robot operates efficiently over the simulated period or fails to achieve the desired objective. In the latter case, poor trajectory tracking at this level would negatively affect subsequent stages, making this a critical prerequisite for proper operation across all remaining benchmark scenarios.\\

\textbf{\textit{Category 2 test - Mid- and low-level control}}\\

This category is built directly on the base configuration, introducing an MPC-TEB control for trajectory tracking while maintaining the same low-level strategy. The objective is to evaluate whether the robot can reach different points in the proposed scenario. To implement this, a model is provided that accounts for the behaviour of the previous category, proposes a minimisation function, and attempts to reach the target position via the route that takes the least time. The analysis focuses on behaviour when reaching the different proposed positions.

The MPC-TEB parameters used are: $v_{\max} = 1.0 $ m/s, $\omega_{\max} = 3.2$ rad/s$^2$, $\lambda_T = 1$, maximum distance between $\mathbf{x}$ and the next predicted state of $0.4$ m, a minimum obstacle distance of $d_{safe} = 0.5 $ m, weighting matrix $Q = 50I$, and input weighting matrix $R = \mathrm{diag}(0.5,\,1.0)$. In this category, the robot starts from the initial pose $(x,y,\theta) = (10.1, 3.0, 0.78)$ and navigates for an approximate duration of 80~s, although the total execution time may vary depending on the convergence of the MPC. The experiment consists of providing four target coordinates within the greenhouse environment (waypoint 1 ($10.2, 13.7$); waypoint 2 ($15.0, 13.7$); waypoint 3 ($18.0, 17.4$)), which the robot must reach while avoiding obstacles and maintaining a minimum safety distance $d_{\text{obst}}$.

The TEB planner incorporates a cost function based on the number of temporal intervals used to estimate the predicted distance to obstacles. Obstacles detected within this horizon are used to construct a local cost map, denoted as \texttt{local\_costmap}. This map enables the algorithm to analyze the surrounding area and plan subsequent motion commands accordingly. The local cost map is built using a 2D LiDAR sensor (Hokuyo URG-04LX), which provides a distance measurement accuracy of $\pm$10~mm within the range of 60~mm to 1~m, and $\pm$1\% of the measured distance for ranges between 1~m and 4~m. As in the first category, no unique solution exists; therefore, three independent trials are conducted, and the mean value and standard deviation of the obtained results are reported.

Figure~\ref{fig:Fig80} illustrates the trajectory to follow by the robot, represented by a green arrow. It is important to note that the plants shown in the image are classified as trajectory-tracking obstacles. Similarly to the previous category, this trajectory is evaluated under two different payload conditions, namely 0 and 70~kg, and four distinct scenarios are considered: i) flat terrain without slope or terrain changes; ii) inclined terrain without terrain changes; iii) flat terrain with terrain changes; and iv) inclined terrain with both slope and terrain changes.

\begin{figure*}[htbp]
\centering
\includegraphics[width=\linewidth]{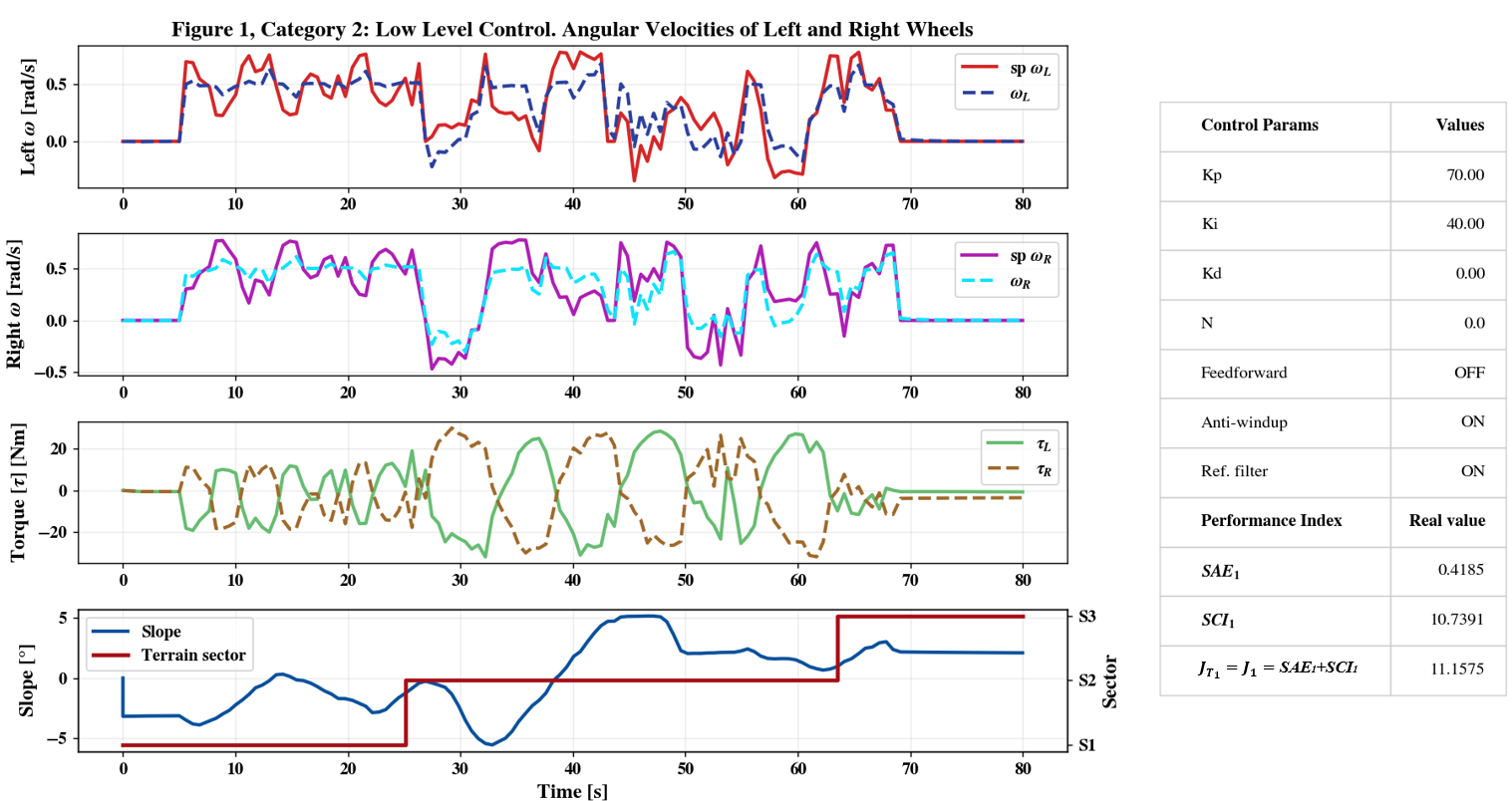}
\caption{Category 2 test, low-level figure, with $m_{pl}=70$, with slope in the terrain and change of sector}
\label{fig:c211}
\end{figure*}

\begin{figure*}[htbp]
\centering
\includegraphics[width=\linewidth]{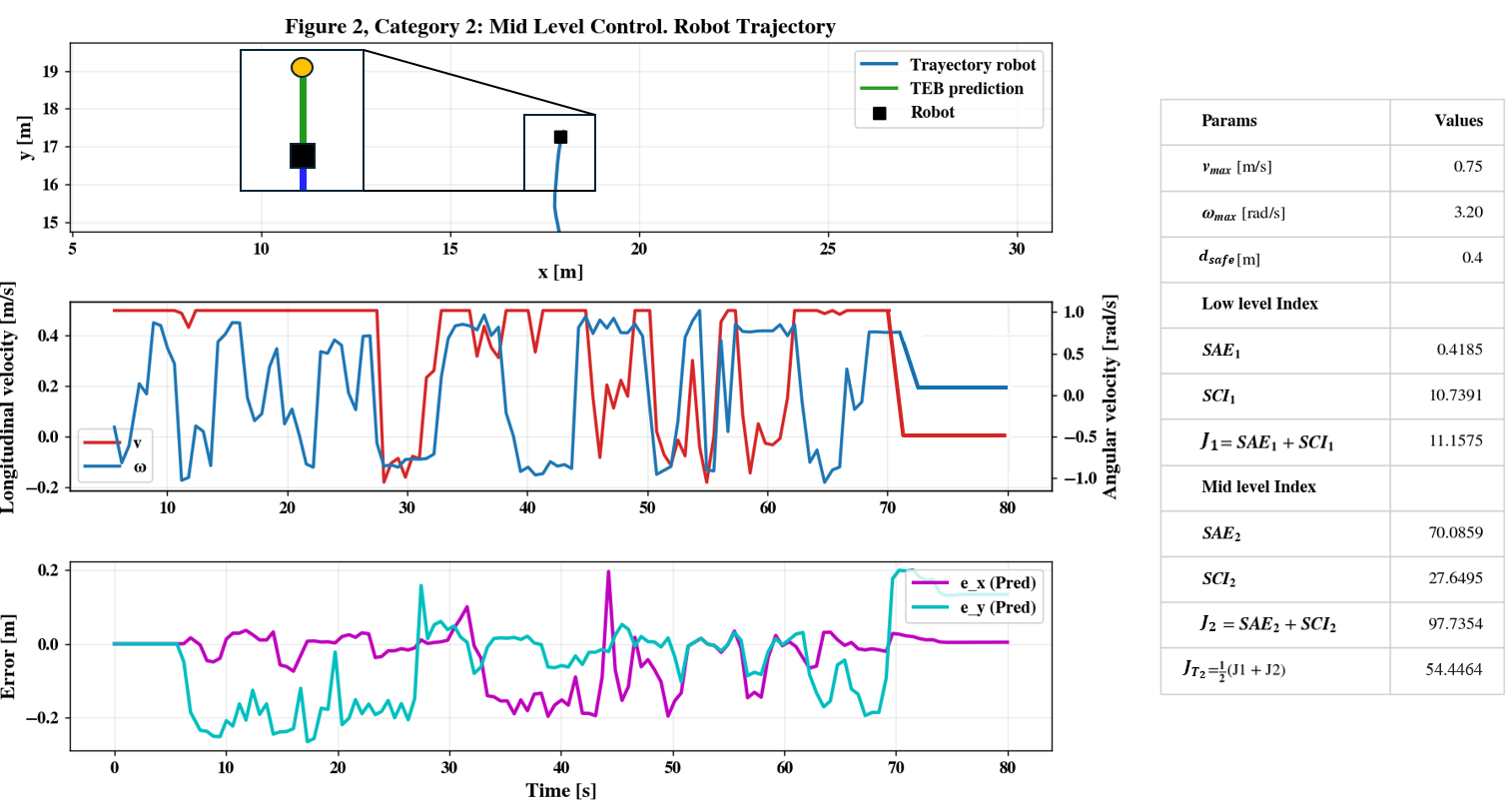}
\caption{Category 2 test, mid-level figure, with $m_{pl}=70$, with slope in the terrain and change of sector. The black box represents the moments before arriving at that area.}
\label{fig:c212}
\end{figure*}

\begin{table}[t]
\centering
\caption{Average SAE and SCI with standard deviation of 3 tests in Category~2 across all terrain and slope conditions.}
\label{tab:MASE_MSIC2}
\begin{adjustbox}{max width=\textwidth}
\begin{tabular}{cc|c|ccc|c}
\toprule
 &  & \multicolumn{1}{c|}{\textbf{Low level}} & \multicolumn{3}{c|}{\textbf{Mid level}} &  \\ 
\cmidrule(lr){3-3} \cmidrule(lr){4-6}
\textbf{Terrain Conditions} & \textbf{Payload [kg]} 
& $J_{1}$  
& $SAE_2$ & $SCI_2$ & $J_{2}$ 
& $J_{T_{2}}$ \\ 
\midrule

\multirow{2}{*}{\shortstack{Slope OFF \\ Terrain OFF}} 
& 0   & 9.9231 $\pm$ 0.4463 & 54.7747 $\pm$ 2.7314 & 23.7730 $\pm$ 1.2385 & 78.5477 $\pm$ 3.9264 & 44.2354 $\pm$ 2.1047 \\
& 70  & 10.3515 $\pm$ 0.5037 & 58.8946 $\pm$ 2.9478 & 24.5896 $\pm$ 1.2796 & 83.4842 $\pm$ 4.0867 & 46.7549 $\pm$ 2.2476 \\ 
\midrule

\multirow{2}{*}{\shortstack{Slope ON \\ Terrain OFF}} 
& 0   & 10.0445 $\pm$ 0.4688 & 55.2447 $\pm$ 2.6215 & 24.0230 $\pm$ 1.1864 & 79.2678 $\pm$ 3.9654 & 44.6562 $\pm$ 2.0976 \\
& 70  & 10.3182 $\pm$ 0.4956 & 58.8946 $\pm$ 2.8739 & 24.7886 $\pm$ 1.3052 & 83.6832 $\pm$ 4.1143 & 46.8327 $\pm$ 2.2381 \\ 
\midrule

\multirow{2}{*}{\shortstack{Slope OFF \\ Terrain ON}} 
& 0   & 9.1898 $\pm$ 0.4296 & 63.0691 $\pm$ 3.1125 & 24.8717 $\pm$ 1.2743 & 87.9409 $\pm$ 4.2961 & 48.3813 $\pm$ 2.3627 \\
& 70  & 10.4285 $\pm$ 0.5019 & 66.6489 $\pm$ 3.2812 & 27.8469 $\pm$ 1.4216 & 84.4958 $\pm$ 4.1369 & 47.2492 $\pm$ 2.1873 \\ 
\midrule

\multirow{2}{*}{\shortstack{Slope ON \\ Terrain ON}} 
& 0   & 11.0549 $\pm$ 0.5281 & 70.0859 $\pm$ 3.4137 & 27.6495 $\pm$ 1.3841 & 97.7354 $\pm$ 4.8873 & 54.2373 $\pm$ 2.6154 \\
& 70  & 12.7284 $\pm$ 0.6114 & 78.5694 $\pm$ 3.7842 & 31.4789 $\pm$ 1.5943 & 110.0483 $\pm$ 5.2671 & 61.3884 $\pm$ 2.9827 \\ 
\bottomrule

\end{tabular}
\end{adjustbox}

\end{table}

The trajectory obtained in this category highlights the impact of employing an MPC-based trajectory tracking strategy. In contrast to Category~1, the wheel velocities exhibit more abrupt variations as a result of the controller convergence process, as illustrated in Fig.~\ref{fig:c211}. These sharper velocity changes are directly reflected in the control input and, consequently, in the KPIs associated with this category. The resulting robot motion is depicted in Fig.~\ref{fig:c212}. In addition, the rectangle with a black border represents the robot's trajectory moments before reaching the target, indicated by the yellow dot (not shown in the legend because this rectangle, along with the yellow dot, was added in post-processing), with the main idea being to show and identify the MPC's convergence. Subplot~1 shows the trajectory followed by the robot from the initial state $\mathbf{x}$ to the goal state $\mathbf{x}_{\text{goal}}$, Subplot~2 presents the control input  $\mathbf{u}$, and Subplot~3 illustrate the trajectory-tracking error over time.

This implementation is reflected in the performance indices for low-level velocity tracking of the robot, which show a noticeable increase due to abrupt changes in angular velocity. These results highlight the impact of introducing a MPC strategy at the subsequent control layer. Both KPIs are reported in Table~\ref{tab:MASE_MSIC2}, where the aforementioned increase observed in Category~1 can be compared with the values obtained for Category~2. As a consequence, the SCI index increases, indicating higher control effort and, therefore, increased energy consumption, which, in turn, reduces the robot’s operational autonomy. This behavior is expected and consistent with the sharper variations in the control signal induced by the changes in the velocity profile.\\

\textbf{\textit{Category 3 test - Hig-, mid- and low-level control}}\\

As previously discussed, this category introduces a path planner into the benchmark by integrating the Lazy~$\theta^*$ algorithm within the simulation framework and implementing a dynamic optimization strategy for trajectory planning based on the $\mathrm{LineOfSight}(n, n')$ criterion. The objective of this strategy is to compute the optimal path between two points while explicitly accounting for an obstacle map obtained from prior environment scanning. The planner relies on a differential-drive robot model, whose parameters are directly derived from those defined in the simulator, thereby preserving the physical and kinematic characteristics considered in the previous benchmark levels. The resulting optimization problem is solved every 2 seconds over the proposed planning horizon, providing a sequence of waypoints that guides the robot from its current state toward the target goal.

Lazy~$\theta^*$ computes traversal costs and penalises high-cost nodes across the map. The selected parameters are: \texttt{how\_many\_corners} = 8 for graph expansion, \texttt{w\_euc\_cost} = 1.0 as the weight applied to path length, and \texttt{w\_traversal\_cost} = 2.0 to penalise high-cost nodes. In this experiment, the robot starts from the initial pose $(x,y,\theta) = (10.1, 3.0, 0.78)$ and navigates for an approximate duration of 80~s. As in Category~2, the total execution time may vary depending on the convergence of the MPC. In this case, a single waypoint located at $(18.0, 17.4)$ is provided. Unlike Category~2, where the robot advances sequentially between waypoints, the robot directly navigates toward the target position following a trajectory planned by the Lazy~$\theta^*$ algorithm.

The global path is generated using a \texttt{global\_costmap} built from a previously constructed map obtained through a SLAM process. This global cost map is continuously compared with the \texttt{local\_costmap} generated by the TEB planner, allowing the computation of a feasible path that avoids obstacles while adapting to local environmental changes. It is worth noting that sensor data are acquired using the same 2D LiDAR sensor as in previous categories, which introduces Gaussian noise. Consequently, no unique solution exists, and three independent trials are also conducted.

Figure~\ref{fig:Fig80} shows the same trajectory as in Category~2, highlighting the difference in waypoint definition. As in the previous categories, this trajectory is evaluated under two payload conditions, namely 0 and 70~kg, and four different scenarios are considered: i) flat terrain without slope or terrain changes; ii) inclined terrain without terrain changes; iii) flat terrain with terrain changes; and iv) inclined terrain with both slope and terrain changes.

In this occasion, the MVSim and RViz2 (Figure \ref{fig:rviz2}) windows also appear, where the map entered, \texttt{local\_costmap} from the MPC and \texttt{global\_costmap} from the scheduler, can be seen. Additionally, the real-time values of the evaluation criteria are displayed, resulting in the computation of the performance index $J_{T_3}$.

\begin{figure}[htbp]
\centering
\includegraphics[width=\linewidth]{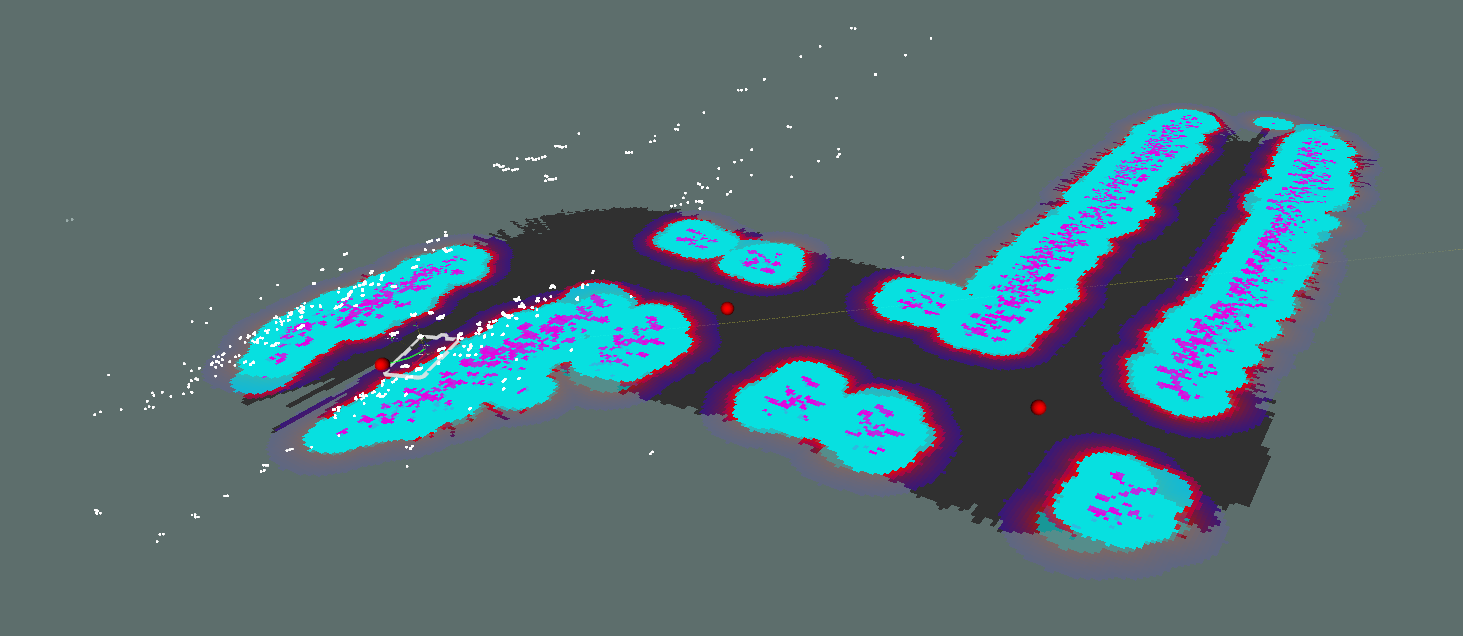}
\caption{Rviz2 for category 3, where the \texttt{local\_costmap} of TEB can be observed}
\label{fig:rviz}
\end{figure}

\begin{figure*}[htbp]
\centering
\includegraphics[width=\linewidth]{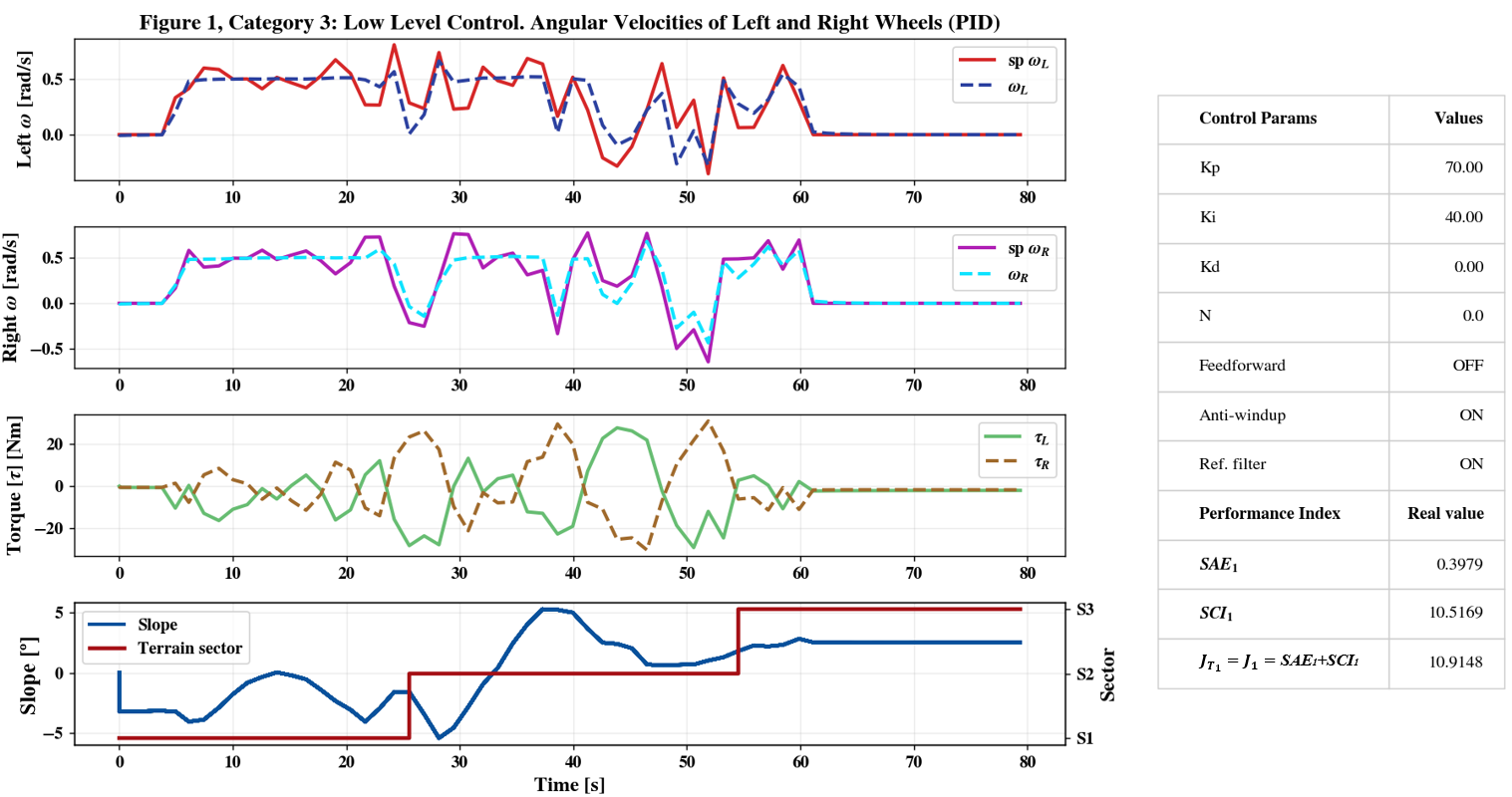}
\caption{Category 3 test, low-level figure, with $m_{pl}=70$, with slope in the terrain and sector change}
\label{fig:c31}
\end{figure*}

\begin{figure*}[htbp]
\centering
\includegraphics[width=\linewidth]{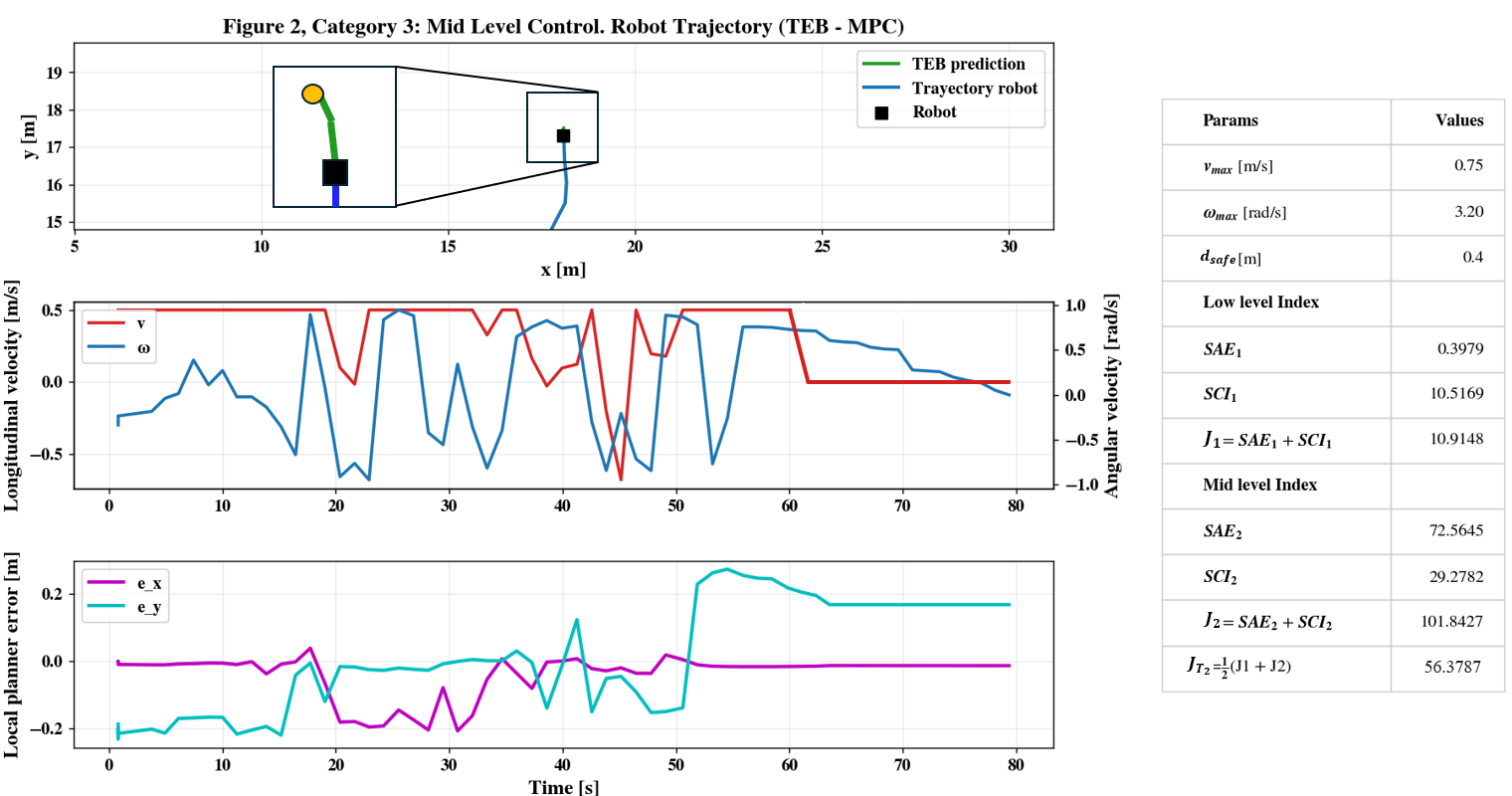}
\caption{Category 3 test, mid-level figure, with $m_{pl}=70$, with slope in the terrain and sector change. The black box represents the moments before arriving at that area.}
\label{fig:c32}
\end{figure*}

\begin{figure*}[htbp]
\centering
\includegraphics[width=\linewidth]{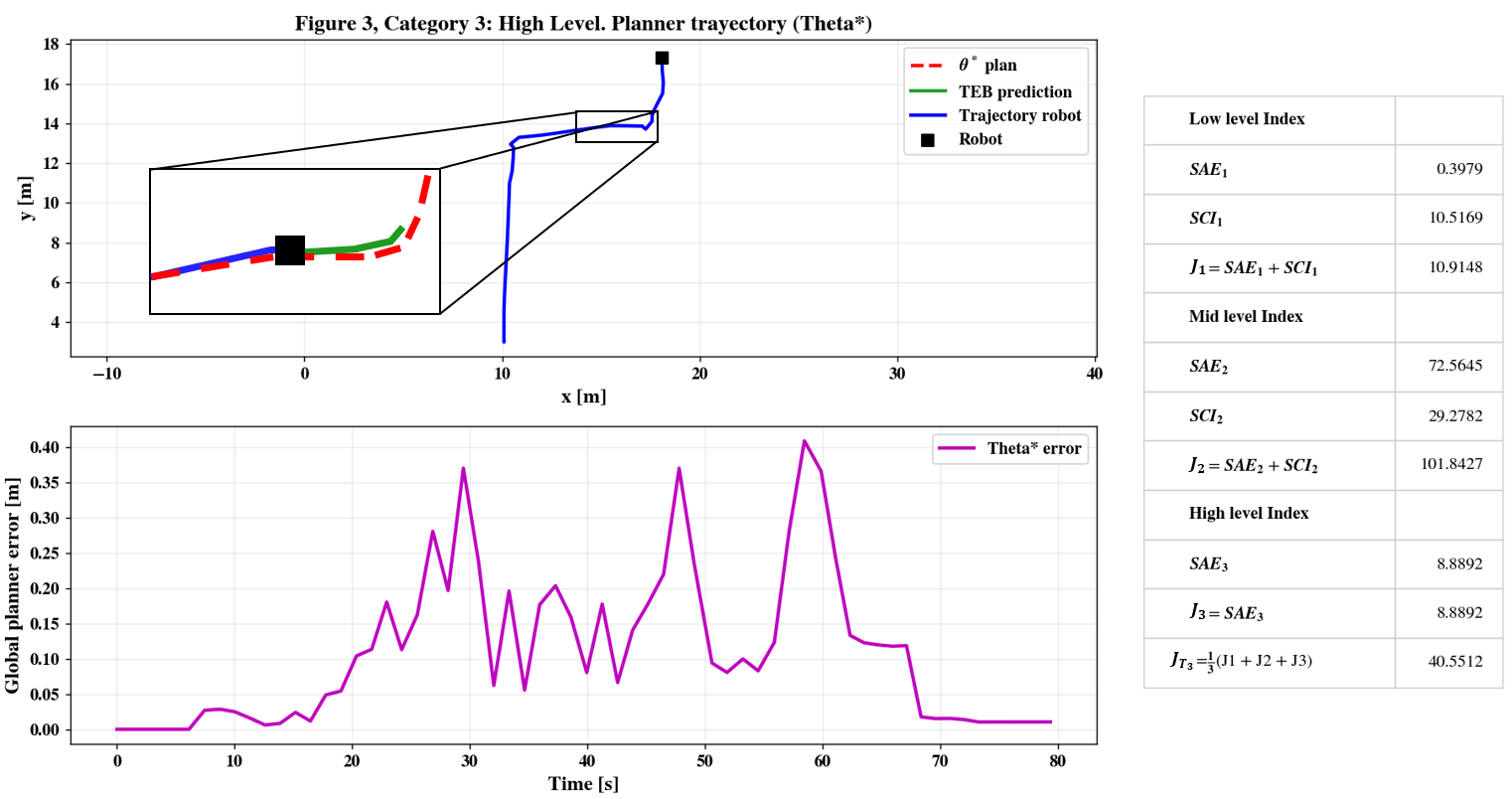}
\caption{Category 3 test, high-level figure, with $m_{pl}=70$, with slope in the terrain and sector changer. The black box represents the moments before arriving at that area.}
\label{fig:c33}
\end{figure*}

\begin{figure}[htbp]
\centering
\includegraphics[width=\linewidth]{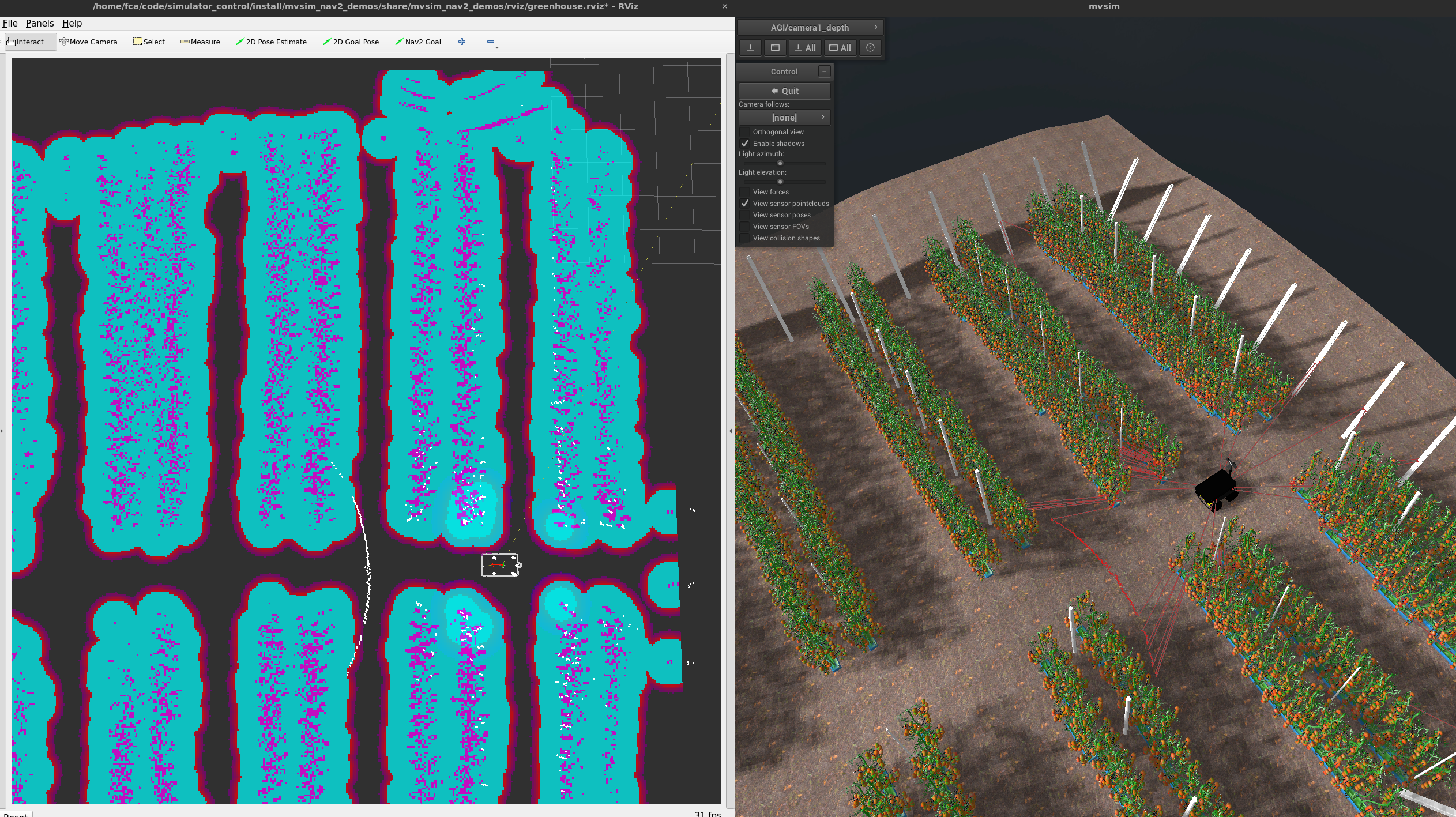}
\caption{Rviz2 and MVSim for category 3, where the \texttt{local\_costmap} of MPC can be observed in lighter shades and the \texttt{global\_costmap} of the planner in darker shades.}
\label{fig:rviz2}
\end{figure}

\begin{table*}[t]
\centering
\caption{Average SAE and SCI with standard deviation of three tests in category 3 across all terrain and slope conditions}
\label{tab:MASE_MSICJ3}
\resizebox{\textwidth}{!}{
\begin{tabular}{c c | c | c | c | c}
\hline
 & & \multicolumn{1}{c|}{\textbf{Low level}} & \multicolumn{1}{c|}{\textbf{Mid level}} & \multicolumn{1}{c|}{\textbf{High level}} &  \multicolumn{1}{c}{\textbf{}} \\ 
\textbf{Terrain Conditions} & \textbf{Payload [kg]} 
& $J_{1}$ & $J_{2}$  & $J_{3}$ =$SAE_3$  & $J_{T_{3}}$\\ \midrule

\multirow{2}{*}{\shortstack[c]{Slope OFF \\ Change of terrain OFF}} 
& 0  & 9.5080 $\pm$ 0.4513 & 88.0518 $\pm$ 4.2365 & 7.5235 $\pm$ 0.3628 & 35.0301 $\pm$ 1.6845 \\
& 70 & 10.2158 $\pm$ 0.4897 & 89.7563 $\pm$ 4.3186 & 7.5824 $\pm$ 0.3691 & 35.8515 $\pm$ 1.7213 \\ \hline

\multirow{2}{*}{\shortstack[c]{Slope ON \\ Change of terrain OFF}} 
& 0  & 11.1867 $\pm$ 0.5234 & 98.4242 $\pm$ 4.9186 & 7.7270 $\pm$ 0.3829 & 39.1172 $\pm$ 1.8927 \\
& 70 & 12.1946 $\pm$ 0.5713 & 101.2563 $\pm$ 5.0842 & 8.1245 $\pm$ 0.4018 & 40.5251 $\pm$ 1.9674 \\ \hline

\multirow{2}{*}{\shortstack[c]{Slope OFF \\ Change of terrain ON}} 
& 0  & 9.3679 $\pm$ 0.4489 & 98.7567 $\pm$ 4.9012 & 6.5145 $\pm$ 0.3197 & 38.2173 $\pm$ 1.8431 \\
& 70 & 9.8563 $\pm$ 0.4726 & 99.7468 $\pm$ 4.9651 & 6.8954 $\pm$ 0.3357 & 38.8328 $\pm$ 1.8794 \\ \hline

\multirow{2}{*}{\shortstack[c]{Slope ON \\ Change of terrain ON}} 
& 0  & 12.0579 ± 1.7737 & 99.7461 ± 3.1870 & 9.0295 ± 0.1231 & 40.5512 ± 1.4941 \\
& 70 & 13.5149 ± 1.9712 & 102.7514 ± 4.2712 & 9.5364 ± 0.1832 & 41.9341 ± 1.5244 \\ \bottomrule

\end{tabular}
}
\end{table*}

Figure~\ref{fig:c31} presents the data associated with Category~1 when influenced by the high-level control layer. Figure~\ref{fig:c32} shows the results corresponding to Category~2, also under the influence of the high-level controller. Finally, Figure~\ref{fig:c33} illustrates the trajectory generated by the planner and followed by the trajectory tracking controller in subplot~1, along with the tracking error in subplot~2. The KPIs associated with the first category are again influenced, though to a lesser extent, as there is little difference in the response profile. As for category 2, the KPIs are also affected, but, as in the previous test, this is not particularly serious, indicating that the medium-level control has followed the trajectory more efficiently than in category 2. Finally, the KPIs in this category indicate that the error between the MPC trajectory and the planner's trajectory has been successfully reduced to low values. These values are reflected in Table \ref{tab:MASE_MSICJ3}.

These values serve as an initial reference for the user to evaluate and compare the performance of the proposed control. Based on this basic configuration, users can adjust the parameters to improve system response, optimise specific criteria, or even completely replace the controllers with their own implementations. In this way, the platform not only allows the nominal behaviour of the robot to be reproduced, but also facilitates experimentation and the design of advanced control strategies adapted to different scenarios and objectives.

\subsection{Evaluating Your Control Law}

This section summarizes how users can employ the proposed control law or evaluate their own controllers. Moreover, a user guide is provided in Appendix section.

\subsubsection{Using the Proposed Controller}

To modify the controller parameters, the user must access the corresponding configuration files provided within the official repository. The directory
\path{"control_params"}, located at
\path{"home/user/ros2_ws/src/robotics_benchmark_greenhouse/benchmark_bringup/benchmark_bringup/control_params/"}, contains the configuration files associated with each experimental category:

\begin{enumerate}
    \item \textbf{\texttt{c1\_pid\_params.yaml}}: File containing the PID controller parameters for Category~1.
    \item \textbf{\texttt{c2\_pid\_params.yaml}}: File containing the MPC controller parameters for Category~2.
    \item \textbf{\texttt{c3\_pid\_params.yaml}}: File containing the Planner controller parameters for Category~3.
\end{enumerate}

Additionally, the directory \path{"home/user/ros2_ws/src/robotics_benchmark_greenhouse/benchmark_bringup/benchmark_bringup/benchmark_bringup/categories/"} provides the Python scripts defining the experimental setups for each category:

\begin{enumerate}
    \item \textbf{\texttt{categorie\_1.py}}: Script specifying the commanded velocities and time intervals for the Category~1 test.
    \item \textbf{\texttt{categorie\_2.py}}: Script defining the waypoints to be optimized by the MPC in Category~2.
    \item \textbf{\texttt{categorie\_3.py}}: Script specifying the target position used by the planner to generate the global path in Category~3.
\end{enumerate}

By modifying the parameters defined in these configuration and script files, users can evaluate alternative control strategies and assess the performance of their own control laws within the proposed benchmarking framework.

\subsubsection{Draft your own control law}

If the user wishes to implement a custom control law in any of the three categories, the benchmark allows the corresponding plugins to be replaced, enabling full customisation of the control architecture. The following guidelines describe how to perform these modifications:

\begin{itemize}

    \item \textit{Category~1}: As previously discussed, the low-level control is implemented within the MVSim simulator. The controller described in Section~4.1 corresponds to a PID-based control function implemented in the simulator source code. This function computes the motor torque $\tau_{m,i}$ from the reference angular velocity $\omega_{\text{ref},i}$, the measured angular velocity $\omega_i$, and the slope-induced torque $\tau_{\text{slope}}$. Given the well-defined input--output structure of this function, users can implement alternative low-level control laws while leveraging the existing simulation infrastructure to compute control performance indices in a consistent manner.

    \item \textit{Category~2}: For this category, as described in previous sections, the mid-level controller is implemented within the Nav2 stack as a controller \textit{plugin}. In addition to the TEB--MPC controller used in this work, Nav2 provides several alternative trajectory-following controllers, such as Dynamic Window Approach (DWB)~\citep{kalliokoski2022hi}, Regulated Pure Pursuit (RPP)~\citep{macenski2023regulated}, and Model Predictive Path Integral (MPPI)~\citep{williams2016aggressive}. To implement a custom mid-level control strategy, users may replace the controller specified in the \texttt{c2\_pid\_params.yaml} configuration file or develop a dedicated Nav2 controller
    plugin. To facilitate this process, a step-by-step tutorial is provided in the Nav2 documentation \footnote{Controller plugin tutorial: \url{https://docs.nav2.org/plugin_tutorials/docs/writing_new_nav2controller_plugin.html}}.

    \item \textit{Category~3}: Similarly to Category~2, the high-level planner is implemented in the Nav2 stack as a planning \textit{plugin}. In addition to the Lazy~$\theta^*$ planner adopted in this benchmark, Nav2 offers alternative global planners, including Navfn and the Smac Planner~\citep{macenski2024open}. To implement a custom high-level planning strategy, users may substitute the planner specified in the \texttt{c3\_pid\_params.yaml} configuration file or develop a new planning plugin. A dedicated step-by-step tutorial is available to support the development of custom Nav2 planner plugins\footnote{Planner plugin tutorial: \url{https://docs.nav2.org/plugin_tutorials/docs/writing_new_nav2planner_plugin.html}}.

\end{itemize}

\section{Conclusions}\label{sec: 8. Conclusion}

This work has presented a comprehensive and reproducible benchmarking framework for the systematic evaluation of mobile robot control strategies in agro-industrial environments, with a particular focus on Mediterranean greenhouses. These environments pose significant challenges due to uneven terrain, variable friction conditions, payload variations, and terrain slopes, all of which have a direct impact on the dynamic behavior and stability of
ground mobile robots. By integrating an accurate 3D geometric representation of the greenhouse, a physics-based simulation environment, and a hierarchical control architecture, the proposed benchmark enables a rigorous analysis of control performance across multiple abstraction levels.

This hierarchical structure evaluates three baseline control architectures, clearly demonstrating the practical utility of the proposed platform and providing a structured view of the benefits associated with increasing levels of control complexity:

\begin{enumerate}
    \item Low-Level Control: The implementation of classical control strategies—including PID control, anti-windup compensation, feedforward action, and reference filtering—tuned using simple model-based rules, enhances the understanding of velocity control in mobile robots. The use of a PI controller for DC motors, combined with these complementary techniques, provides effective, robust control even in the presence of disturbances. At this level, the user can define three distinct disturbance scenarios, enabling a systematic evaluation of the proposed control laws using performance indices (KPIs). The results demonstrate that basic feedback mechanisms, supported by first-principles modeling, are essential for achieving smooth, stable regulation of low-level control variables. 

    \item Mid-Level Control: Implementing Model Predictive Control (MPC) strategies for trajectory tracking highlights the potential of directly embedding optimization into the control architecture. By dynamically optimizing the trajectory between two points, the TEB-MPC approach successfully accounts for constraints that maintain a safe distance from obstacles. Although this strategy increases the control effort at the low level—and consequently the energy consumption—it enables reliable waypoint-to-waypoint navigation in the presence of both static and dynamic obstacles. This clearly demonstrates the strong potential of optimization-based control strategies for trajectory tracking. Furthermore, the considered disturbance scenarios—namely, payload variation, terrain type, and terrain slope—faithfully reproduce real-world operating conditions commonly encountered in agricultural robotics, providing a realistic and challenging testbed for robust, adaptive, and predictive control approaches.

    \item High-Level Control: The integration of path planning algorithms significantly improves high-level control performance through dynamic optimization. The planned path is updated every second, ensuring successful goal achievement while avoiding collisions with obstacles identified in a previously loaded static map. This implementation reduces control effort, resulting in lower energy consumption and greater robot autonomy compared to the mid-level control category. These results highlight the importance of coordinated planning and control for achieving efficient, scalable autonomous navigation.
\end{enumerate}


A key contribution of this work is the definition of standardized performance metrics of KPIs, including SAE and SCI indexes, and combined cost functions. These metrics enable quantitative, objective, and reproducible comparisons between different control strategies under identical experimental conditions. The statistical evaluation based on repeated trials further strengthens the reliability of the obtained results and mitigates the influence of stochastic sensor noise and environmental variability.

Moreover, the adoption of a plugin-based architecture, compatible with widely used robotic software frameworks, facilitates the seamless integration of user-defined controllers and planners. This design choice significantly enhances the extensibility, adaptability, and long-term scalability of the benchmark, promoting its use as an open reference platform for the research community. As a result, the proposed framework not only supports fair comparative evaluation but also accelerates the development and validation of novel control and planning algorithms tailored to agricultural applications.

Future work will focus on extending the benchmark toward real-world experimental validation, incorporating additional sensing modalities, and considering more complex agricultural tasks such as manipulation, perception-driven navigation, and multi-robot coordination. Ultimately, this benchmark aims to bridge the gap between simulation-based research and real agro-industrial deployments, fostering knowledge transfer between the robotics research community and the agricultural sector. This positions it as a key tool for facilitating the methodological process and closing the gap between control theory and the actual operation of robots.

\section*{Acknowledgments} 
The first author, Fernando Cañadas-Aránega, holds an FPI grant (PRE2022-102415) from the Spanish Ministry of Science, Innovation, and Universities, and this work was partially supported by Agencia Estatal de Investigación (AEI) under the Project PID2022-139187OB-I00.
\bibliographystyle{cas-model2-names}

\bibliography{cas-refs}

\section*{Author Note}

During the preparation of this work, the authors used \textit{ChatGPT (OpenAI)} as an assistance tool for text generation and technical editing. After using this tool, the authors thoroughly reviewed, modified, and validated the content, and they take full responsibility for the final version of the manuscript.

\section*{Appendix A. User guide}

This section provides the necessary information for users to operate the benchmark and to develop their own controllers for each of the defined categories.
\begin{itemize}
    \item \textbf{Benchmark Installation Guide} 
\end{itemize}

The MVSim simulator has been developed on Ubuntu 22.04 (Jammy) using the \texttt{ROS~2 Humble} framework. However, it is also compatible with Ubuntu~24.04~(Jammy) and any other supported \texttt{ROS~2 DISTRO}. Previously, you must install ROS2 Humble \footnote{ROS2 Humble repository: \url{https://docs.ros.org/en/humble/Installation.html}}, the Nav2 navigation package \footnote{Nav2 stack: \url{https://docs.nav2.org/}}, and \texttt{mrpt} \footnote{mrpt library: \url{https://docs.mrpt.org/reference/latest/}} library. Once ROS~2 is installed, the user may clone the official Benchmark Control\footnote{Official benchmark repository: \url{https://github.com/FerCanAra/robotics_benchmark_greenhouse}} repository from GitHub. This repository contains this tree:

\begin{enumerate}
    \item \texttt{benchmark\_control}: This is the package that houses the control scheme configuration files, as well as the generation of graphs, routes, maps, sensor definitions, robots, worlds, and general configuration files.
    
    \item \texttt{mvsim}: This is the core of the simulator, developed by the University of Almeria itself \citep{blanco2023multivehicle}.
    
    \item \texttt{teb\_local\_planner}: This package implements a local navigator plugin to the base\_local\_planner of the 2D navigation stack. The underlying method called Timed Elastic Band locally optimizes the robot's trajectory with respect to trajectory execution time, separation from obstacles and compliance with kinodynamic constraints at runtime.
    
    \item \texttt{costmap\_converter}: used by TEB to convert environmental information into the \texttt{local\_costmap} format required by \texttt{nav2}.
\end{enumerate}


To install the complete simulator, execute:

\begin{lstlisting}[language=bash]
mkdir -p ~/ros2_ws/src
cd ~/ros2_ws/src
git clone --recurse-submodules https://github.com/FerCa
nAra/robotics_benchmark_greenhouse.git
cd robotics_benchmark_greenhouse
git submodule update --init --recursive
cd ../../..
colcon build --symlink install -DCMAKE_BUILD_TYPE=Relea
se -DCMAKE_EXPORT_COMPILE_COMMANDS=ON
source install/setup.bash
\end{lstlisting}

With these simple commands, you will have the complete simulator at your disposal. You can launch the default demo using the following command:

\begin{lstlisting}[language=bash]
ros2 launch mvsim benchmark_control.launch.py
\end{lstlisting}

\begin{itemize}
    \item \textbf{Simulation Configuration}
\end{itemize}

With the environment installed, the user may configure the simulator according to their requirements. The following parameters must be set:

\begin{enumerate}
    \item \textit{Category selection}: specify the benchmark category using \texttt{category:=1} (valid values: 1, 2 or 3).
    \item \textit{Payload selection}: choose a mass between $0\,\mathrm{kg}$ and $70\,\mathrm{kg}$.
    \item \textit{Perturbation selection}: activate terrain slope, terrain change or both using the boolean parameters \\[2pt]
    \texttt{terrain\_slope:=true} \quad \texttt{change\_terrain:=true}.
\end{enumerate}

With these settings, the simulator loads the parameters described previously to compute the values summarised in Table 1. A complete launch example is:

\begin{lstlisting}[language=bash]
ros2 launch mvsim benchmark_control.launch.py \
    category:=1 \
    payload:=70 \
    terrain_slope:=true \
    change_terrain:=true
\end{lstlisting}

The user can select the combination they want, obtaining a simulation tailored to their selection. Once the simulation is complete, a .csv file is generated in an automatically generated path where the simulator was launched \path{"pwd/result/category_X/"} with the following format \texttt{yyyy\_mm\_dd\_hh\_mm\_ss\\.csv} where all the data for the category is stored. With this information, the user can perform post-processing to carry out the actions they deem appropriate.

\newpage

\begin{table*}[t]
\centering
\footnotesize
\caption{Robot model, control variables, disturbances, planning parameters, and performance indices}
\label{tab:variables}
\nonumber
\begin{adjustbox}{max width=\textwidth}
\begin{tabular}{lll|lll}
\toprule

\multicolumn{6}{c}{\textbf{Robot and Physical Parameters}} \\
\midrule
Symbol & Description & Unit & Symbol & Description & Unit \\
\midrule
$\mathbf{x}$ & Robot state vector $(x,y,\theta)$ & -- &
$x$ & Robot position in x & m \\
$y$ & Robot position in y & m &
$\theta$ & Robot orientation & rad \\
$\mathbf{u}$ & Control input $(v,\omega)$ & -- &
$v$ & Longitudinal robot velocity & m/s \\
$\omega$ & Angular robot velocity & rad/s &
$m_v$ & Chassis mass & kg \\
$m_w$ & Wheel mass & kg &
$m_{robot}$ & Total robot mass & kg \\
$n_W$ & Number of wheels & -- &
$r$ & Wheel radius & m \\
$L_w$ & Wheel separation & m &
$I_{yy}$ & Robot inertia & kg\,m$^2$ \\
$b$ & Motor viscous friction & N\,m\,s\,rad$^{-1}$ &
$J$ & Motor inertia & kg\,m$^2$ \\
$\tau_{m,i}$ & Motor input torque & N\,m &
$i$ & Motor index, $\in \{R,L\}$ & -- \\
$K_\tau$ & Motor torque constant & N\,m &
$C_D$ & Motor damping coefficient & -- \\
$\omega_{ref}$ & Reference angular velocity & rad/s &
$m_{pw}$ & Total mass distributed per wheel & kg \\
$\omega_{i}$ & Angular velocity per motor & rad/s &  &  & \\

\midrule
\multicolumn{6}{c}{\textbf{Disturbances and External Effects}} \\
\midrule
$F_{t,i}$ & Total traction force per motor & N &
$F_{long,i}$ & Longitudinal force per motor & N \\
$F_{r_{max}}$ & Maximum friction force & N &
$F_{lat,i}$ & Latitudinal force per motor & N \\
$g$ & Gravity acceleration & m/s$^2$ &
$a_{wy}$ & Lateral acceleration & rad/s$^2$ \\
$\alpha_i$ & Angular velocity variation coefficient & -- &
$m_{pl}$ & Payload mass & kg \\
$p$ & Payload indicator & -- &
$\phi$ & Terrain slope & $\degree$ \\
$C_{rr}$ & Rolling resistance coefficient & -- &
$s$ & Terrain type indicator & -- \\
$\mu$ & Friction coefficient & -- & & & \\

\midrule
\multicolumn{6}{c}{\textbf{Low-Level Control (Velocity Control)}} \\
\midrule
$K_m$ & Motor gain & N\,m\,s\,rad$^{-1}$ &
$\tau_{motor}$ & Motor time constant & s \\
$\tau_{PID,i}$ & PID output torque & N\,m &
$e_{\omega,i}$ & Motor velocity error & rad/s \\
$K_P$ & Proportional gain (PID) & N\,m\,s &
$K_I$ & Integral gain (PID) & N\,m \\
$K_D$ & Derivative gain (PID) &  N\,m\,s &
$N$ & Derivative filter constant & -- \\
$\tau_{d,i}$ & Derivative torque & N\,m &
$\tau_{max}$ & Maximum saturation torque & N\,m \\
$\tau_{mff,i}$ & Feedforward torque & N\,m &
$\tau_{m,i_{sat}}$ & Saturated motor torque & N\,m \\
$K_{aw}$ & Anti-windup gain & s$^{-1}$ &
$I$ & Integral action & N\,m \\
$\tau_f$ & Reference filter time constant & s &
$n_f$ & Reference filter order & -- \\
$\omega_{fil,i}$ & Filtered angular velocity & rad/s &
$F_{Crr}$ & Rolling resistance force & N \\
$F_g$ & Gravitational force & N &
$F_{slope}$ & Slope-induced force & N \\
$\tau_{D,i}$ & Damping torque & Nm &
$\tau_{slope}$ & Slope-induced torque & Nm \\
$K_s$ & Static gain relating $\Delta\omega_i$ to $\Delta\tau_{slope}$ & rad\,s$^{-1}$\, N$^{-1}$\, m$^{-1}$ &
$K_{ff}$ & Static Feedforward gain & -- \\
$\tau_{slope_{real}}$ & Real torque applied to the robot & N\, m & &  &  \\

\midrule
\multicolumn{6}{c}{\textbf{Mid-Level Control (Trajectory Tracking)}} \\

\midrule
$k$ & Discrete time index & -- &
$\mathcal{H}$ & Polytopic approx. robot’s footprint  & -- \\
$\mathbb{Y}$ & Robot workspace & -- &
$Z$ & Set the status restrictions and inputs & -- \\
$\mathcal{O}$ & Polygonal obstacles & -- &
$n_o$ & Number of obstacles & -- \\
$j$ & Obstacle index & -- &
$\delta_{\text{obst}}$ & Distance to obstacles & m \\
$P$ & Prediction horizon & -- &
$\mathbf{x}_{ref}$ & Reference state vector & -- \\
$\lambda_T$ & TEB temporal weight & -- &
$Q$ & State weighting matrix & -- \\
$R$ & Control weighting matrix & -- &
$d_{safe}$ & Safety distance & m \\
$v_{max}$ & Maximum linear velocity & m/s &
$\omega_{max}$ & Maximum angular velocity & rad/s \\
$r_{robot}$ & Robot footprint radius & m &
$r_{\mathcal{O}_j}$ & Obstacle radius & m \\
$\mathbf{p_{o_{O_j}}}$ & Obstacle postion & -- &
$x_{teb}$ & MPC converged $x$ position & m \\
$y_{teb}$ & MPC converged $y$ position & m & & & \\

\midrule
\multicolumn{6}{c}{\textbf{High-Level (Planner)}} \\

\midrule
$n$ & Number of nodes & -- &
$n_{goal}$ & Goal node & -- \\
$\mathbf{u_{ref}}$ & Reference control input & -- &
$c_{o}$ & Maximum number of corners & -- \\
$n_{max}$ & Maximum number of nodes & -- &
$w_{traversal}$ & Traversal cost weight & -- \\
$\mathbb{x}_{plan}$ & Planned state sequence & -- $w_{euc}$ & t & -- & \\

\midrule
\multicolumn{6}{c}{\textbf{Index Performance}} \\

\midrule
$n_c$ & Category index & -- &
$J_{n_c}$ & Category performance index & -- \\
$SAE_{n_c}$ & Sum of Absolute Error & -- &
$SCI_{n_c}$ & Control Control Signal & -- \\
$J_{T_{n_c}}$ & Total cost per category & -- & & & \\

\bottomrule
\end{tabular}
\end{adjustbox}
\end{table*}




\end{document}